\documentclass{article}
\usepackage{spconf,amsmath,graphicx,booktabs,url}
\usepackage{newtxtext,newtxmath}
\usepackage[hidelinks]{hyperref}
\usepackage{balance}
\let\standardthebibliography\thebibliography
\renewcommand{\thebibliography}[1]{%
  \standardthebibliography{#1}\setlength{\itemsep}{-0.2pt}}

\title{GGUF-METADATA PREDICTION OF SINGLE-SEQUENCE LLAMA.CPP THROUGHPUT ACROSS THREE SYSTEMS}
\name{Xinyu Qiu\textsuperscript{1}, Chuhong Xu\textsuperscript{2},
  Bo Su\textsuperscript{3}, Ziyao Chen\textsuperscript{4},
  Ruiyang Xu\textsuperscript{1}, Shimeng Dai\textsuperscript{5}}
\address{%
  {\small \textsuperscript{1}Northeastern University \quad
    \textsuperscript{2}Sofia University \quad \textsuperscript{3}Indiana University}\\
  {\small \textsuperscript{4}University of California, San Diego \quad
    \textsuperscript{5}Michigan State University}\\
  {\small \nolinkurl{qiu.xiny@northeastern.edu},
    \nolinkurl{chuhong.xu@sofia.edu}, \nolinkurl{subo@iu.edu}}\\
  {\small \nolinkurl{cziyao@ucsd.edu},
    \nolinkurl{xu.r@northeastern.edu}, \nolinkurl{daishime@msu.edu}}}

\AddToHookNext{shipout/foreground}{%
  \put(54,-24){\makebox(0,0)[lt]{%
    \parbox{504pt}{\normalfont\fontsize{9}{10}\selectfont
      This work has been submitted to the IEEE for possible publication. Copyright may be transferred without notice, after which this version may no longer be accessible.}}}}
\begin{document}
\ninept
\setlength{\emergencystretch}{1em}
\maketitle

\begin{abstract}
We predict single-sequence model throughput from GGUF metadata using
roofline-shaped predictors with quantization-specific scale factors fitted on
reference models.  The scored cohort comprises 318 phase--depth measurements
from 53 host--file configurations on two Apple M4 Max systems and an NVIDIA RTX
5080.  On host-specific held-out sets of four, five, and two configurations, an
active-parameter decode model obtains 13.1\%, 14.4\%, and 36.1\% mean absolute
percentage error (MAPE), versus 49.4\%, 55.3\%, and 51.9\% when charging total
parameters.  Leave-one-host-out coefficients fitted on the other two systems
yield 11.6\%, 16.8\%, and 36.0\% test MAPE.  A low-bit model ladder changes
ordering across runtime stacks.  The P2 prefill baseline gives 18.7\%, 22.2\%,
and 108.2\% test MAPE.  GGUF structure helps on all three systems, but fitted
efficiencies are not universal.
\end{abstract}


\section{Introduction}
\label{sec:intro}

Choosing a local language model often requires benchmarking several
multi-gigabyte GGUFs.  A predictor based on stored metadata would reduce
candidate executions relative to a lookup table; remote header-only acquisition
is outside our scope.  Three effects complicate a raw memory roofline
\cite{williams2009roofline}: quantized kernels attain different fractions
of peak bandwidth, mixture-of-experts (MoE) models store many more weights than
they activate per token, and hybrid architectures allocate key--value (KV) state differently
in global, sliding-window, and recurrent layers.

We ask which parts of such a predictor transfer across model families and
hardware.  Our first contribution is a host-adapted decode model whose target
features---file size, tensor shapes, expert routing, and per-layer attention
metadata---are available before inference.  We evaluate a frozen family split
on two unified-memory Apple systems and a discrete NVIDIA GPU.  Second, we
isolate the benefit of activated-parameter and per-layer KV accounting.  Third,
we report transfer limits rather than hiding them in a pooled score: one-term
coefficients only partly transfer, low-bit behavior varies by system, and
the tested prefill model does not replicate on RTX.

LLMCompass explores accelerator design \cite{zhang2024llmcompass}.  Vidur
combines profiling with serving simulation \cite{agrawal2024vidur}.
RooflineBench and LIMINAL apply hardware-aware analysis to LLM inference
\cite{bi2026rooflinebench,davies2025liminal}.
MoE-CAP argues that sparse utilization should be measured over activated rather
than stored parameters \cite{jiang2025moecap}; we test the predictive effect of
that accounting.  Benazir and Lin characterize quantization and memory behavior
on Apple Silicon \cite{benazir2025apple}.  Our narrower estimator targets
unmeasured single-sequence GGUF families from metadata plus a small reference
set, followed by an explicit cross-system falsification test.  Measurements use \texttt{llama-bench} from
\texttt{llama.cpp} \cite{gerganov2023llamacpp}; GGUF parsing follows its format
specification \cite{ggufspec}.

\begin{figure*}[!t]
  \centering
  \includegraphics[width=\textwidth]{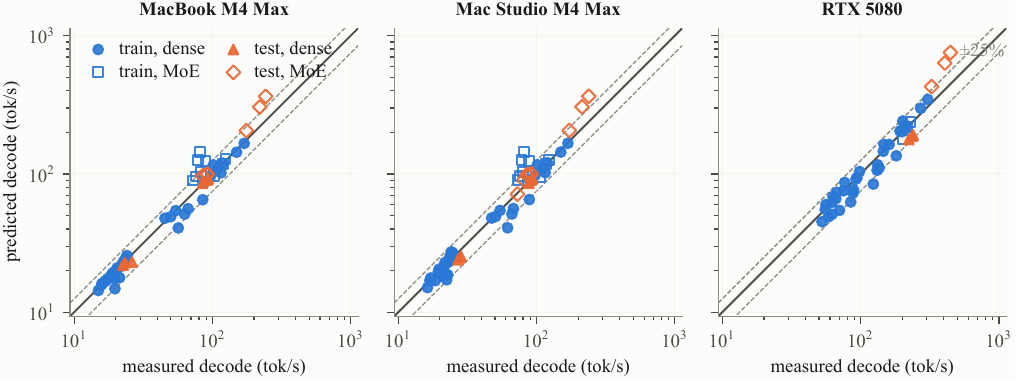}
  \caption{B2 decode predictions, separated by host.  The 159 points comprise
  57 MacBook, 60 Studio, and 42 RTX observations, with 126 training and 33 held-out; dashed
  lines mark $\pm25\%$.  Open markers are MoE and orange markers are held-out.}
  \label{fig:accuracy}
\end{figure*}

\section{Predictor}
\label{sec:model}

Let measured $T_d(c)$ and $T_p(c)$ denote decode and prefill throughput.  Let
$S$ be GGUF file size, $P$ and $P_a$ total and activated parameter counts,
$B_h$ measured bandwidth on host $h$, and $c$ existing-prefix depth.  We model
bytes per generated token and decode tokens per second as
\begin{align}
 D(c)&=S\frac{P_a}{P}+K(c), &
 \widehat T_d(c)&=\eta_{h,q}\frac{B_h}{D(c)}, \label{eq:decode}\\[-2mm]
 K(c)&=\sum_i2h_i d_i\min(c,w_i)b_{\rm KV}+S_{\rm state}. \label{eq:kv}
\end{align}
Here $q$ is quantization format, $h_i$ and $d_i$ are a layer's KV-head count
and head dimension, $w_i$ is its window ($\infty$ for global attention), and
$b_{\rm KV}=2$ bytes per cached scalar for the fixed F16 cache, and
$S_{\rm state}$ is fixed recurrent-state traffic.  Recurrent layers set $h_i=0$.
Tensor dimensions and expert-routing fields
estimate $P_a$; if an expert bank has $P_e$ weights, $E$
experts, and $k$ selected experts, then
$P_a=P-P_e+(k/E)P_e$.  Dense models have $P_a=P$.

For each $(h,q)$, $\eta_{h,q}$ is the median of
$T_dD/B_h$ over training rows; an unseen format uses the host-wide median.
This ratio-space fit prevents the fastest small
models from dominating least squares.  It is an empirical scale factor, not
literal utilization: $S P_a/P$ assumes file-wide average bytes per parameter,
and both file bytes and the copy benchmark are traffic proxies, so
$\eta$ may exceed one.  B0 is the unfitted total-size roofline
$B_h/(S+K)$; B1 fits $\eta_{h,q}$ but still charges total weights; B2 uses
Eqs.~\eqref{eq:decode}--\eqref{eq:kv}.  Comparing B1 with B2 isolates
active-weight accounting while holding the fitting rule fixed.  Because $B_h$
cancels after same-host fitting, it affects B0 and cross-host substitution, not
target-fitted B1/B2 accuracy; $F_h$ analogously cancels for P1/P2.

We also test the deliberately simple zero-prefix prefill model
\begin{equation}
 \widehat T_p=\phi_{h,q}\frac{F_h}{2P_a}, \label{eq:prefill}
\end{equation}
where $F_h$ is calibrated matrix throughput.  For each depth-zero training row,
we back-solve $r=T_p(2P_a)/F_h$.  P1 sets $\phi_h$ to the median $r$ for host
$h$; P2 uses the median within each host--quantization group.  An unseen
quantization falls back to the median of that host's fitted group coefficients.
Neither model contains an existing-prefix term, so greater depths are scope
tests, not part of the fit.

\begin{figure*}[!t]
  \centering
  \includegraphics[width=\textwidth]{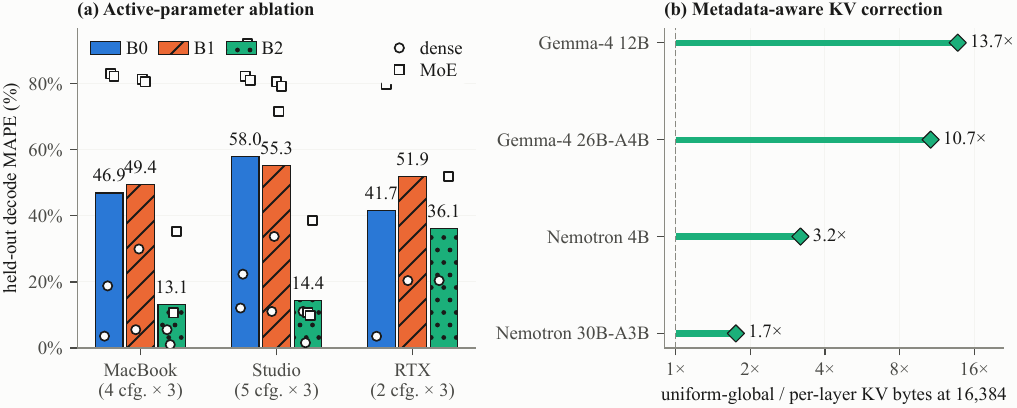}
  \caption{Structural corrections.  (a) Bars are host-specific held-out MAPE;
  overlaid points are per-configuration means across three depths and expose
  the four-, five-, and two-configuration test sizes.  (b) From GGUF metadata, treating every layer
  as global overstates 16K KV bytes by as much as $13.7\times$.}
  \label{fig:mechanisms}
\end{figure*}

\begin{figure*}[!t]
  \centering
  \includegraphics[width=\textwidth]{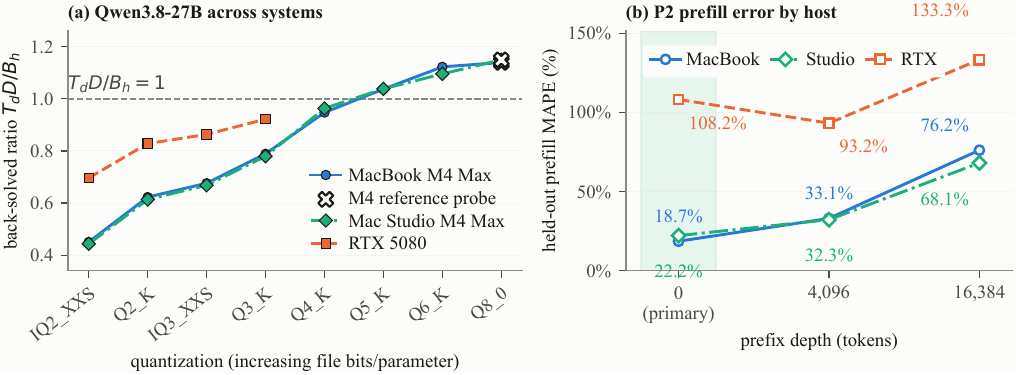}
  \caption{Transfer diagnostics.  (a) The back-solved depth-zero ratio
  $T_dD/B_h$ for one Qwen family varies across systems; RTX covers the first
  four formats.  The crossed Apple Q8 points are historical reference probes,
  shown for ladder context but excluded from fitting and scoring.  (b) P2
  prefill error is host-specific, remains poor on RTX, and rises by 16K.}
  \label{fig:scope}
\end{figure*}

\section{Experimental method}
\label{sec:method}

The hosts are a 64-GB MacBook Pro M4 Max and a 128-GB Mac Studio M4 Max, each
using 12 performance-core threads, plus an RTX 5080 with 16,302~MiB reported
VRAM, CUDA, and 16 threads.  Their device-copy/FP16 matrix calibrations are
380.1~GB/s and 13.95~TFLOP/s, 396.7~GB/s and 15.00~TFLOP/s, and 801.1~GB/s and
118.83~TFLOP/s, respectively.  The Studio binary is from the hash-recorded
official b10794 archive.  The managed RTX directory is labeled b10794, but its
version query returned usage text; the legacy MacBook revision was not recovered.

The shared manifest has 23 GGUF files.  The MacBook produced 22 successful
grids; after three pre-specified reference-probe exclusions, it contributes 15
training and four held-out configurations.  Studio completed all 23 grids and
contributes 15 and five after the same exclusions.  RTX covers 14 matching files,
split 12 and two.  The 53 scored configurations span 22 files; each held-out set
contains dense and MoE families absent from training, though RTX remains a small
replication.  Results are therefore host-separated.  Splits are fixed by base
family before fitting.  The 63.39-GB gpt-oss-120B file produced two prompt-batch
failures on MacBook, not a demonstrated out-of-memory failure, and a complete
grid on Studio.  The supplement lists every file, source URL, revision, and byte
size; source organizations are cited here \cite{unsloth_models,lmstudio_models,
ggmlorg_models,google_models,bartowski_models}.

All reported throughput is single-sequence.  Each configuration runs in a
fresh process.  At prefix depths 0, 4096, and 16384, \texttt{llama-bench}
reports separate 512-token prompt-processing and 128-token generation
microbenchmarks, not one end-to-end request.  Runs use flash attention, F16 K/V,
five repetitions, and 45~s settling.  The requested GPU-layer count is 99.  Physical
RTX residency was not instrumented; Windows shared-memory spill therefore
cannot be excluded for the largest files.  The append-only data are reduced by
an atomic selector that prefers complete protocol rows and then lower decode
coefficient of variation (CV)
and pre-run load, without inspecting prediction residuals.

The stability gate applies to decode.  Across all selected rows, median decode
CV is 0.83\%, 0.42\%, and 0.43\% on MacBook, Studio, and RTX; maxima are 3.04\%,
2.30\%, and 1.65\%.  Prefill was not gated: 8/66, 0/69, and 29/42 rows exceed 3\%,
with maxima 4.84\%, 0.96\%, and 39.2\%.  We retain them and give a median-sample
sensitivity below.  Each decode configuration contributes its three depths.

\section{Results}
\label{sec:results}

\begin{table}[t]
\centering
\caption{Host-separated MAPE (\%).  ``Cfg.'' is the number of model--format
configurations; decode includes three depths per configuration, while P1/P2
are scored at depth zero.}
\label{tab:error}
\setlength{\tabcolsep}{3.0pt}
\begin{tabular}{@{}llrrrrr@{}}
\toprule
Host & Split (cfg.) & B0 & B1 & B2 & P1 & P2\\
\midrule
MacBook & train (15) & 33.6 & 15.8 & \textbf{10.4} & 6.8 & \textbf{4.1}\\
        & test (4)   & 46.9 & 49.4 & \textbf{13.1} & 22.7 & \textbf{18.7}\\
Studio & train (15) & 33.0 & 17.8 & \textbf{12.4} & 7.0 & \textbf{4.3}\\
       & test (5)   & 58.0 & 55.3 & \textbf{14.4} & 28.0 & \textbf{22.2}\\
RTX 5080 & train (12) & 20.3 & \textbf{9.8} & 10.1 & 14.5 & \textbf{5.9}\\
         & test (2)   & 41.7 & 51.9 & \textbf{36.1} & 124.5 & 108.2\\
\bottomrule
\end{tabular}
\end{table}

\subsection{Host-adapted decode}

Figure~\ref{fig:accuracy} and Table~\ref{tab:error} show the primary result.
B2 train/test MAPE is 10.4/13.1\% on MacBook, 12.4/14.4\% on Studio, and
10.1/36.1\% on RTX.  RTX test median/maximum error is 25.8/69.0\%.  On the exact
two-file intersection shared by all hosts, MAPE is 18.2\%, 20.1\%, and 36.1\%,
respectively.  RTX remains the smallest replication, but B2 improves on B1 for
every held-out host.

Active-weight accounting substantially improves the held-out aggregate
(Fig.~\ref{fig:mechanisms}a): B1 errors of 49.4\%, 55.3\%, and 51.9\% fall to
13.1\%, 14.4\%, and 36.1\%.  Capacity is controlled by stored parameters,
whereas per-token weight traffic is closer to activated parameters.  Separately,
Eq.~\eqref{eq:kv} avoids the
$10.7$--$13.7\times$ cache overestimates produced by a uniform-global formula
for the two Gemma-4 examples (Fig.~\ref{fig:mechanisms}b).  After refitting each
variant, per-layer rather than uniform-global KV lowers held-out MAPE from
21.2\% to 13.1\% on MacBook, 21.0\% to 14.4\% on Studio, and 37.9\% to 36.1\% on RTX.

\begin{figure*}[!t]
  \centering
  \includegraphics[width=\textwidth]{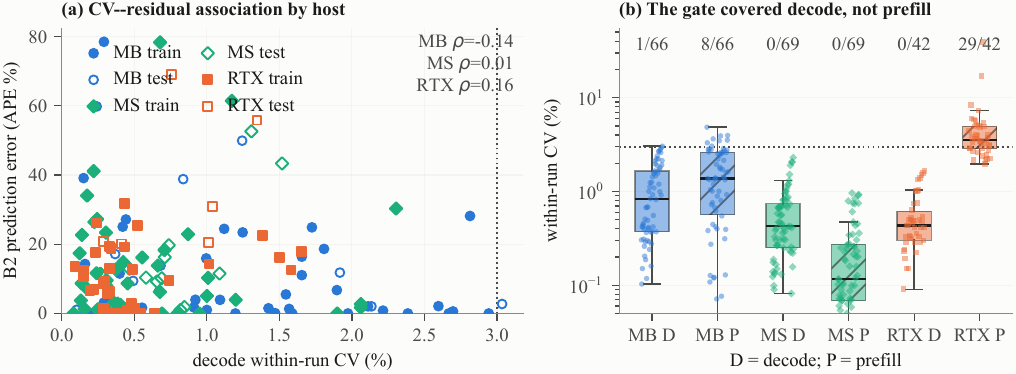}
  \caption{Measurement diagnostics.  (a) Host-separated decode CV shows no obvious descriptive association
  with B2 error across 159 scored rows; the dotted line is the 3\% target.
  (b) Row-level CV distributions use all 354 selected rows, including 36 Apple
  reference-probe rows; labels count rows above the target.  The gate covered
  decode (D), not prefill (P).}
  \label{fig:quality}
\end{figure*}

\subsection{Cross-host transfer is incomplete}

Separate fitting should not be confused with hardware generalization.  In a
leave-one-host-out diagnostic, B2 coefficients are fitted independently on the
other two machines, given equal source-host weight, and paired with the target's
calibrated bandwidth.  All-target MAPE is 14.5\%, 15.4\%, and 20.8\% for 57
MacBook, 60 Studio, and 42 RTX rows; target-fitted B2 gives 11.0\%, 12.9\%, and
13.8\%.  Restricted to target test families, transfer gives 11.6\%, 16.8\%, and
36.0\%, close to target-fitted 13.1\%, 14.4\%, and 36.1\%.  This test is easier
than simultaneous model-and-host holdout because most target files occur on a
source host.

The 14 three-way matched configurations span eight base families and weight each
format equally.  Across 22 MacBook--Studio matches, Studio/MacBook geometric-mean
zero-prefix throughput is $1.02\times$ for both decode and prefill.  RTX/MacBook
is $2.15\times$ for decode, near the $2.11\times$ copy-bandwidth ratio, and
$7.21\times$ for prefill versus an $8.52\times$ compute ratio.  Configuration-level
decode ratios span $1.76$--$3.26\times$, ruling out one universal multiplier.

The near-family Qwen3.8-27B ladder varies across the three host/runtime stacks
(Fig.~\ref{fig:scope}a).  IQ2 has 1.55\% fewer parameters and one fewer layer.
On both Apple hosts, Q2 is only 2.2\% faster than IQ2 at depth zero, whereas on
RTX, IQ2 is 13.7\% faster.  Across the three conventional Q4/Q8 pairs, Q4 is
$1.32$--$1.43\times$ faster on MacBook, $1.29$--$1.41\times$ on Studio, and
$1.37$--$1.49\times$ on RTX, while its
prefill rate is usually slightly lower.  File compression alone therefore does
not specify kernel efficiency.

\subsection{Prefill is a negative result}

P1/P2 held-out zero-prefix MAPE is 22.7/18.7\% on MacBook, 28.0/22.2\% on
Studio, and 124.5/108.2\% on RTX (Fig.~\ref{fig:scope}b).
The RTX result is not solely the 39.2\%-CV Ling cell:
replacing every cell mean by its five-sample median and refitting still gives
85.5\%.  At 16K, where Eq.~\eqref{eq:prefill} omits prefix-dependent work, test
error rises to 76.2\% on MacBook, 68.1\% on Studio, and 133.3\% on RTX.  Most
host--quantization groups contain only one training configuration, so the
per-format coefficients are weakly identified and P2's low 4.1--5.9\% training
errors are partly resubstitution.  We consequently treat prefill as
unsupported under this protocol, not as a successful second regime.

\subsection{Protocol and robustness checks}

Decode behaves more consistently than its largest residuals.  Every scored
configuration slows monotonically with context.  B2 held-out MAPE falls from
17.9\% to 8.6\% between depth zero and 16K on MacBook, 18.1\% to 9.7\% on Studio,
and 44.5\% to 25.6\% on RTX.  Across all 159 scored decode rows, within-run CV
shows no obvious descriptive association with residual error (pooled Spearman
$\rho=-0.05$; host values are $-0.14$, $0.01$, and $0.16$;
Fig.~\ref{fig:quality}), and within-run scatter is much smaller than the largest
residuals.

Correlated depth rows and absent Studio/RTX between-run studies limit noise
attribution.  The supplement reports a rejected output-projection extension.

\section{Limitations and conclusion}
\label{sec:conclusion}

The RTX held-out set has only two Q4\_K\_M files and no independent between-run
repeatability campaign; acquisition spans two sessions.  Quantization formats
often have one training family, IQ2's Qwen file has 1.55\% fewer parameters
than the rest of that ladder, and execution order was not randomized.  The
largest RTX analytical working set exceeds reported VRAM before graph buffers,
so requested full layer offload is not proof of physical residency.  No GPU
clock, power, or spill telemetry was recorded.  Studio files were SHA-verified
against repository LFS identifiers; equivalent immutable captures are
unavailable for the legacy hosts.  These limitations preclude a universal runtime
accuracy claim.  All error summaries are descriptive, not population confidence
estimates.  The $SP_a/P$ proxy also scales shared tensors with expert tensors
and assumes balanced $k/E$ routing; tensor-level active-byte validation was
unavailable.  No learning curve establishes how many reference configurations
are sufficient.

Within that boundary, active-weight and metadata-aware KV corrections reduce
held-out decode error in all three cohorts; source-host-balanced coefficients
remain close after target bandwidth calibration.  Host/runtime-stack differences
in low-bit behavior argue against a universal coefficient; differing runtime
versions and IQ2's slightly different architecture remain confounds.  Next work
should record residency and broaden the cohort.

\clearpage
\setlength{\emergencystretch}{1.5em}
\Urlmuskip=0mu plus 1mu
\balance
\bibliographystyle{IEEEbib}
\bibliography{references}

\end{document}


\begin{center}
{\Large\bfseries Companion Results Appendix: GGUF Throughput Prediction\par}
\vspace{4pt}
{\normalsize Protocol-complete 3-host evidence for the ICASSP manuscript\par}
\end{center}
\textbf{Status.} This is a locally generated companion appendix. It does not claim publication, archival acceptance, or independent replication.
\section{Scope and cohort}
The source manifest contains 23 GGUF files. Across 3 hosts (MacBook Pro M4 Max, Mac Studio M4 Max, and RTX 5080), the scored data contain 22 unique files and 53 host--file configurations (42 training and 11 held out). Each configuration has one decode and one prefill observation at each of the 3 measured context depths, for 159 decode and 159 prefill rows. The final protocol selector retains 354 observations from 23 unique files in total: 318 scored rows and 36 reference-probe rows excluded from prediction fits and scores. Probe exclusion is host-specific, so a file used as a probe on one host can be scored on another.
All exported rows carry the declared protocol fields. The retry decision, however, uses the worst within-run \emph{decode} CV in an invocation. Prefill rows inherit that invocation's protocol metadata but are not gated on their own CV. This distinction matters on RTX 5080, where the noisiest prefill row has 39.19\% CV even though the associated decode-gate CV is 1.35\%.
\begin{center}
\small
\begin{tabular}{llrrrr}
\toprule
Host & Backend & Train cfg. & Test cfg. & Scored decode & Scored prefill\\
\midrule
MacBook Pro M4 Max & BLAS,MTL & 15 & 4 & 57 & 57 \\
Mac Studio M4 Max & MTL,BLAS & 15 & 5 & 60 & 60 \\
RTX 5080 & CUDA & 12 & 2 & 42 & 42 \\
\bottomrule
\end{tabular}
\end{center}
\begin{center}
\small
\begin{tabular}{lrrrrrr}
\toprule
Host & Retried cells & Gate CV med. & Gate CV max & PP CV med. & PP CV max & PP $>3\%$\\
\midrule
MacBook Pro M4 Max & 3 & 1.43 & 3.04 & 1.41 & 4.84 & 7/57 \\
Mac Studio M4 Max & 1 & 0.88 & 2.30 & 0.14 & 0.96 & 0/60 \\
RTX 5080 & 3 & 0.88 & 1.65 & 3.56 & 39.19 & 29/42 \\
\bottomrule
\end{tabular}
\end{center}
Attempted configurations with no successful row are MacBook Pro M4 Max: \nolinkurl{gpt-oss-120b-MXFP4.gguf}. On the MacBook Pro M4 Max, two raw attempts of the one gpt-oss-120B configuration report \texttt{failed to decode prompt batch, res=-3}; these are prompt-batch failures, not demonstrated model-load or out-of-memory failures. Host-specific calibration probes are MacBook Pro M4 Max: \nolinkurl{Qwen3.8-27B-Q8_0.gguf}, \nolinkurl{Qwen3.5-9B-Q8_0.gguf}, \nolinkurl{Qwen3.5-4B-Q8_0.gguf}; Mac Studio M4 Max: \nolinkurl{Qwen3.8-27B-Q8_0.gguf}, \nolinkurl{Qwen3.5-9B-Q8_0.gguf}, \nolinkurl{Qwen3.5-4B-Q8_0.gguf}; RTX 5080: none.
\subsection{Cohort accounting}
Raw records, superseded legacy rows, repeated protocol attempts, selected observations, and scoring exclusions are separated below. A protocol duplicate is a successful protocol row removed by the atomic host--model--phase--depth selector; it is not an additional scored observation.
\begin{center}
\small
\setlength{\tabcolsep}{2.8pt}
\begin{tabular}{lrrrrrrrr}
\toprule
Host & Raw & Failed & Legacy ok & Protocol ok & Dup. removed & Selected & Probes & Scored\\
\midrule
MacBook Pro M4 Max & 278 & 2 & 132 & 144 & 12 & 132 & 18 & 114 \\
Mac Studio M4 Max & 138 & 0 & 0 & 138 & 0 & 138 & 18 & 120 \\
RTX 5080 & 84 & 0 & 0 & 84 & 0 & 84 & 0 & 84 \\
\bottomrule
\end{tabular}
\end{center}
\subsection{Excluded reference-probe observations}
These rows complete the 354-observation selected cohort and make the descriptive matched-host and quantization comparisons reproducible. They are reported only as measurements, never as prediction errors.
\begingroup
\small
\setlength{\tabcolsep}{3.5pt}
\renewcommand{\arraystretch}{1.08}
\begin{longtable}{@{}lllrrrrr@{}}
\caption{Selected protocol-complete reference-probe observations excluded from prediction fitting and scoring. Measured throughput and SD are in tokens/s; IDs resolve through Table~\ref{tab:model-id}.}\label{tab:probe-observations}\\
\toprule
Host & ID & Phase & Depth & Measured & SD & CV \% & Attempts \\
\midrule
\endfirsthead
\caption[]{Selected protocol-complete reference-probe observations excluded from prediction fitting and scoring. Measured throughput and SD are in tokens/s; IDs resolve through Table~\ref{tab:model-id}. (continued)}\\
\toprule
Host & ID & Phase & Depth & Measured & SD & CV \% & Attempts \\
\midrule
\endhead
\midrule
\multicolumn{8}{r}{Continued on next page}\\
\endfoot
\bottomrule
\endlastfoot
MacBook Pro M4 Max & M10 & decode & 0 & 77.275 & 0.201 & 0.260 & 1 \\
MacBook Pro M4 Max & M10 & decode & 4,096 & 74.687 & 0.153 & 0.204 & 1 \\
MacBook Pro M4 Max & M10 & decode & 16,384 & 67.270 & 0.341 & 0.507 & 1 \\
MacBook Pro M4 Max & M10 & prefill & 0 & 1553.086 & 1.682 & 0.108 & 1 \\
MacBook Pro M4 Max & M10 & prefill & 4,096 & 1416.098 & 1.799 & 0.127 & 1 \\
MacBook Pro M4 Max & M10 & prefill & 16,384 & 992.403 & 10.398 & 1.048 & 1 \\
MacBook Pro M4 Max & M12 & decode & 0 & 46.499 & 0.380 & 0.818 & 2 \\
MacBook Pro M4 Max & M12 & decode & 4,096 & 45.340 & 0.169 & 0.372 & 2 \\
MacBook Pro M4 Max & M12 & decode & 16,384 & 42.119 & 0.938 & 2.226 & 2 \\
MacBook Pro M4 Max & M12 & prefill & 0 & 865.397 & 0.661 & 0.076 & 2 \\
MacBook Pro M4 Max & M12 & prefill & 4,096 & 784.470 & 4.299 & 0.548 & 2 \\
MacBook Pro M4 Max & M12 & prefill & 16,384 & 560.769 & 14.295 & 2.549 & 2 \\
MacBook Pro M4 Max & M14 & decode & 0 & 14.902 & 0.054 & 0.365 & 1 \\
MacBook Pro M4 Max & M14 & decode & 4,096 & 14.433 & 0.017 & 0.118 & 1 \\
MacBook Pro M4 Max & M14 & decode & 16,384 & 13.575 & 0.092 & 0.680 & 1 \\
MacBook Pro M4 Max & M14 & prefill & 0 & 238.511 & 7.257 & 3.043 & 1 \\
MacBook Pro M4 Max & M14 & prefill & 4,096 & 201.115 & 2.027 & 1.008 & 1 \\
MacBook Pro M4 Max & M14 & prefill & 16,384 & 161.599 & 1.227 & 0.759 & 1 \\
Mac Studio M4 Max & M10 & decode & 0 & 78.751 & 0.546 & 0.693 & 1 \\
Mac Studio M4 Max & M10 & decode & 4,096 & 76.142 & 0.100 & 0.131 & 1 \\
Mac Studio M4 Max & M10 & decode & 16,384 & 70.660 & 0.398 & 0.563 & 1 \\
Mac Studio M4 Max & M10 & prefill & 0 & 1567.538 & 4.181 & 0.267 & 1 \\
Mac Studio M4 Max & M10 & prefill & 4,096 & 1430.704 & 3.482 & 0.243 & 1 \\
Mac Studio M4 Max & M10 & prefill & 16,384 & 1129.638 & 0.567 & 0.050 & 1 \\
Mac Studio M4 Max & M12 & decode & 0 & 48.365 & 0.091 & 0.188 & 1 \\
Mac Studio M4 Max & M12 & decode & 4,096 & 47.519 & 0.047 & 0.099 & 1 \\
Mac Studio M4 Max & M12 & decode & 16,384 & 45.097 & 0.041 & 0.090 & 1 \\
Mac Studio M4 Max & M12 & prefill & 0 & 880.266 & 0.530 & 0.060 & 1 \\
Mac Studio M4 Max & M12 & prefill & 4,096 & 834.386 & 0.599 & 0.072 & 1 \\
Mac Studio M4 Max & M12 & prefill & 16,384 & 722.102 & 0.565 & 0.078 & 1 \\
Mac Studio M4 Max & M14 & decode & 0 & 15.697 & 0.063 & 0.403 & 1 \\
Mac Studio M4 Max & M14 & decode & 4,096 & 15.505 & 0.041 & 0.266 & 1 \\
Mac Studio M4 Max & M14 & decode & 16,384 & 14.931 & 0.022 & 0.150 & 1 \\
Mac Studio M4 Max & M14 & prefill & 0 & 255.725 & 0.175 & 0.068 & 1 \\
Mac Studio M4 Max & M14 & prefill & 4,096 & 244.576 & 0.168 & 0.069 & 1 \\
Mac Studio M4 Max & M14 & prefill & 16,384 & 215.717 & 0.047 & 0.022 & 1 \\
\end{longtable}
\endgroup
\paragraph{Probe-exclusion sensitivity.}
For each host with calibration probes, B2 is refit after adding those files back. The original-train column evaluates that refit only on non-probe training rows; test rows remain held out.
\begin{center}
\small
\setlength{\tabcolsep}{3.0pt}
\begin{tabular}{lrrrrrrr}
\toprule
Host & Probe cfg. & Aug. train $n$ & Aug. MAPE & Original $n$ & Original MAPE & Test $n$ & Test MAPE\\
\midrule
MacBook Pro M4 Max & 3 & 54 & 10.46 & 45 & 11.06 & 12 & 13.11 \\
Mac Studio M4 Max & 3 & 54 & 12.65 & 45 & 13.39 & 15 & 14.37 \\
\bottomrule
\end{tabular}
\end{center}
\section{Aggregate prediction results}
Results are separated by host because fitted efficiency factors are host-specific; a pooled row would weight the unequal host cohorts and is not a cross-system score.
\begin{center}
\small
\begin{tabular}{lllrrrr}
\toprule
Host & Predictor & Split & $n$ & MAPE (\%) & Median (\%) & P90 / max (\%)\\
\midrule
MacBook Pro M4 Max & B0 unfitted (eta=1, total) & train & 45 & 33.57 & 19.37 & 76.47 / 123.27 \\
MacBook Pro M4 Max & B0 unfitted (eta=1, total) & test & 12 & 46.89 & 49.97 & 83.45 / 84.22 \\
MacBook Pro M4 Max & B1 fitted, total params & train & 45 & 15.80 & 3.20 & 70.29 / 76.77 \\
MacBook Pro M4 Max & B1 fitted, total params & test & 12 & 49.36 & 54.66 & 81.90 / 82.74 \\
MacBook Pro M4 Max & B2 fitted, active params (ours) & train & 45 & 10.39 & 2.12 & 26.32 / 78.49 \\
MacBook Pro M4 Max & B2 fitted, active params (ours) & test & 12 & 13.11 & 8.98 & 36.65 / 49.92 \\
Mac Studio M4 Max & B0 unfitted (eta=1, total) & train & 45 & 32.95 & 19.40 & 75.52 / 125.41 \\
Mac Studio M4 Max & B0 unfitted (eta=1, total) & test & 15 & 57.98 & 81.60 & 92.02 / 92.67 \\
Mac Studio M4 Max & B1 fitted, total params & train & 45 & 17.76 & 7.49 & 68.32 / 75.84 \\
Mac Studio M4 Max & B1 fitted, total params & test & 15 & 55.25 & 72.08 & 80.95 / 81.72 \\
Mac Studio M4 Max & B2 fitted, active params (ours) & train & 45 & 12.40 & 7.49 & 29.09 / 78.32 \\
Mac Studio M4 Max & B2 fitted, active params (ours) & test & 15 & 14.37 & 11.30 & 33.89 / 52.62 \\
RTX 5080 & B0 unfitted (eta=1, total) & train & 36 & 20.27 & 14.47 & 44.15 / 71.19 \\
RTX 5080 & B0 unfitted (eta=1, total) & test & 6 & 41.65 & 40.39 & 81.22 / 81.95 \\
RTX 5080 & B1 fitted, total params & train & 36 & 9.85 & 8.88 & 20.95 / 31.73 \\
RTX 5080 & B1 fitted, total params & test & 6 & 51.85 & 50.81 & 84.50 / 85.11 \\
RTX 5080 & B2 fitted, active params (ours) & train & 36 & 10.12 & 9.34 & 20.95 / 31.73 \\
RTX 5080 & B2 fitted, active params (ours) & test & 6 & 36.15 & 25.79 & 62.41 / 68.96 \\
\midrule
MacBook Pro M4 Max & P1 eta\_p per host & train & 15 & 6.81 & 5.36 & 16.58 / 33.40 \\
MacBook Pro M4 Max & P1 eta\_p per host & test & 4 & 22.68 & 19.16 & 44.95 / 51.25 \\
MacBook Pro M4 Max & P2 eta\_p per host x quant & train & 15 & 4.13 & 0.00 & 15.06 / 25.99 \\
MacBook Pro M4 Max & P2 eta\_p per host x quant & test & 4 & 18.68 & 14.90 & 36.90 / 42.85 \\
Mac Studio M4 Max & P1 eta\_p per host & train & 15 & 7.00 & 2.17 & 18.89 / 36.84 \\
Mac Studio M4 Max & P1 eta\_p per host & test & 5 & 28.04 & 30.15 & 46.54 / 47.19 \\
Mac Studio M4 Max & P2 eta\_p per host x quant & train & 15 & 4.27 & 0.00 & 16.14 / 26.29 \\
Mac Studio M4 Max & P2 eta\_p per host x quant & test & 5 & 22.23 & 20.11 & 40.03 / 42.81 \\
RTX 5080 & P1 eta\_p per host & train & 12 & 14.46 & 9.15 & 31.38 / 52.16 \\
RTX 5080 & P1 eta\_p per host & test & 2 & 124.51 & 124.51 & 207.24 / 227.93 \\
RTX 5080 & P2 eta\_p per host x quant & train & 12 & 5.86 & 0.00 & 21.31 / 25.36 \\
RTX 5080 & P2 eta\_p per host x quant & test & 2 & 108.18 & 108.18 & 184.89 / 204.07 \\
\bottomrule
\end{tabular}
\end{center}
Decode summaries use all three context depths. Prefill headline rows use only depth 0, because the stated prefill model has no existing-prefix term; deeper rows are scope diagnostics below.
\subsection{Leave-one-host-out transfer}
B2 fits median per-quantization coefficients on all non-target training hosts and substitutes the untouched target host's bandwidth. The target set contains all eligible rows on that host; because most files also occur on at least one source host, this isolates host/runtime transfer rather than a simultaneous model-and-hardware holdout. A target format absent from all source training rows uses the source-wide median fallback. The rejected two-term rows are the absolute-time output-projection extension.
\begin{center}
\small
\begin{tabular}{llrrrrr}
\toprule
Variant & Target host & Source $n$ & Target $n$ & MAPE & Median & Max\\
\midrule
B2: all target rows & MacBook Pro M4 Max & 81 & 57 & 14.47 & 8.02 & 73.61 \\
B2: all target rows & Mac Studio M4 Max & 81 & 60 & 15.42 & 10.78 & 84.30 \\
B2: all target rows & RTX 5080 & 90 & 42 & 20.80 & 18.18 & 68.13 \\
B2: target test rows & MacBook Pro M4 Max & 81 & 12 & 11.59 & 6.00 & 45.82 \\
B2: target test rows & Mac Studio M4 Max & 81 & 15 & 16.76 & 14.36 & 57.74 \\
B2: target test rows & RTX 5080 & 90 & 6 & 35.97 & 25.66 & 68.13 \\
rejected two-term & MacBook Pro M4 Max & 81 & 57 & 17.88 & 12.10 & 75.66 \\
rejected two-term & Mac Studio M4 Max & 81 & 60 & 16.93 & 9.50 & 66.06 \\
rejected two-term & RTX 5080 & 90 & 42 & 149.33 & 132.38 & 501.57 \\
\bottomrule
\end{tabular}
\end{center}
For context, target-fitted B2 gives all-row MAPE of 10.96\% on MacBook Pro M4 Max; 12.89\% on Mac Studio M4 Max; 13.84\% on RTX 5080. Unlike the untouched-target transfer rows, these comparators mix training rows used to fit the target coefficients with held-out rows.
\begin{center}
\small
\begin{tabular}{lllrrrr}
\toprule
Host & Split & Architecture & $n$ & B0 & B1 & B2\\
\midrule
MacBook Pro M4 Max & train & dense & 36 & 23.21 & 7.14 & 5.66 \\
MacBook Pro M4 Max & train & MoE & 9 & 74.98 & 50.45 & 29.29 \\
MacBook Pro M4 Max & test & dense & 6 & 11.18 & 17.75 & 3.28 \\
MacBook Pro M4 Max & test & MoE & 6 & 82.61 & 80.97 & 22.95 \\
Mac Studio M4 Max & train & dense & 36 & 22.74 & 9.91 & 7.89 \\
Mac Studio M4 Max & train & MoE & 9 & 73.79 & 49.15 & 30.45 \\
Mac Studio M4 Max & test & dense & 6 & 17.21 & 22.40 & 6.31 \\
Mac Studio M4 Max & test & MoE & 9 & 85.16 & 77.15 & 19.74 \\
RTX 5080 & train & dense & 33 & 15.73 & 10.57 & 10.57 \\
RTX 5080 & train & MoE & 3 & 70.28 & 1.95 & 5.18 \\
RTX 5080 & test & dense & 3 & 3.54 & 20.40 & 20.40 \\
RTX 5080 & test & MoE & 3 & 79.77 & 83.31 & 51.90 \\
\bottomrule
\end{tabular}
\end{center}
\subsection{Per-layer KV correction}
With each variant refit on training rows, replacing a uniform-global KV denominator by the frozen per-layer metadata produces:
\begin{center}
\small
\begin{tabular}{llrrr}
\toprule
Host & Split & $n$ & Uniform KV MAPE (\%) & Per-layer KV MAPE (\%)\\
\midrule
MacBook Pro M4 Max & train & 45 & 12.68 & 10.39 \\
MacBook Pro M4 Max & test & 12 & 21.23 & 13.11 \\
Mac Studio M4 Max & train & 45 & 14.68 & 12.40 \\
Mac Studio M4 Max & test & 15 & 20.96 & 14.37 \\
RTX 5080 & train & 36 & 12.37 & 10.12 \\
RTX 5080 & test & 6 & 37.87 & 36.15 \\
\bottomrule
\end{tabular}
\end{center}
\subsection{Decode context behavior}
All 53 scored host--file configurations slow monotonically from 0 through 16,384 tokens. The normalized rate is each model's throughput divided by its own depth-0 throughput.
\begin{center}
\small
\begin{tabular}{lrrrrrr}
\toprule
Host & Depth & Train $n$ & Test $n$ & Train B2 MAPE & Test B2 MAPE & Median normalized rate\\
\midrule
MacBook Pro M4 Max & 0 & 15 & 4 & 11.30 & 17.86 & 1.000 \\
MacBook Pro M4 Max & 4,096 & 15 & 4 & 7.59 & 12.83 & 0.945 \\
MacBook Pro M4 Max & 16,384 & 15 & 4 & 12.27 & 8.63 & 0.837 \\
Mac Studio M4 Max & 0 & 15 & 5 & 13.60 & 18.13 & 1.000 \\
Mac Studio M4 Max & 4,096 & 15 & 5 & 8.07 & 15.28 & 0.976 \\
Mac Studio M4 Max & 16,384 & 15 & 5 & 15.54 & 9.69 & 0.915 \\
RTX 5080 & 0 & 12 & 2 & 10.19 & 44.52 & 1.000 \\
RTX 5080 & 4,096 & 12 & 2 & 4.49 & 38.28 & 0.985 \\
RTX 5080 & 16,384 & 12 & 2 & 15.67 & 25.64 & 0.922 \\
\bottomrule
\end{tabular}
\end{center}
\subsection{Prefill scope diagnostic}
Depth 0 is the primary prefill fit and score. The same fitted coefficients are applied unchanged at larger existing-prefix depths. These errors are prediction errors, distinct from within-run prefill variability; results remain separated by host.
\begin{center}
\small
\begin{tabular}{llrrrrrr}
\toprule
Host & Split & Depth & $n$ & P1 MAPE & P2 MAPE & P2 median & P2 max\\
\midrule
MacBook Pro M4 Max & train & 0 & 15 & 6.81 & 4.13 & 0.00 & 25.99 \\
MacBook Pro M4 Max & train & 4,096 & 15 & 22.35 & 20.83 & 17.58 & 49.82 \\
MacBook Pro M4 Max & train & 16,384 & 15 & 66.94 & 64.13 & 51.02 & 150.97 \\
MacBook Pro M4 Max & test & 0 & 4 & 22.68 & 18.68 & 14.90 & 42.85 \\
MacBook Pro M4 Max & test & 4,096 & 4 & 40.98 & 33.11 & 29.60 & 55.36 \\
MacBook Pro M4 Max & test & 16,384 & 4 & 86.59 & 76.18 & 73.78 & 132.42 \\
Mac Studio M4 Max & train & 0 & 15 & 7.00 & 4.27 & 0.00 & 26.29 \\
Mac Studio M4 Max & train & 4,096 & 15 & 15.96 & 13.66 & 8.81 & 40.52 \\
Mac Studio M4 Max & train & 16,384 & 15 & 47.15 & 42.84 & 18.25 & 115.02 \\
Mac Studio M4 Max & test & 0 & 5 & 28.04 & 22.23 & 20.11 & 42.81 \\
Mac Studio M4 Max & test & 4,096 & 5 & 38.87 & 32.32 & 32.35 & 57.15 \\
Mac Studio M4 Max & test & 16,384 & 5 & 78.65 & 68.11 & 63.93 & 126.96 \\
RTX 5080 & train & 0 & 12 & 14.46 & 5.86 & 0.00 & 25.36 \\
RTX 5080 & train & 4,096 & 12 & 19.26 & 12.35 & 8.23 & 32.74 \\
RTX 5080 & train & 16,384 & 12 & 36.13 & 28.56 & 18.19 & 72.26 \\
RTX 5080 & test & 0 & 2 & 124.51 & 108.18 & 108.18 & 204.07 \\
RTX 5080 & test & 4,096 & 2 & 108.39 & 93.23 & 93.23 & 167.57 \\
RTX 5080 & test & 16,384 & 2 & 151.64 & 133.34 & 133.34 & 227.36 \\
\bottomrule
\end{tabular}
\end{center}
\begin{figure}[t]
\centering
\includegraphics[width=\textwidth]{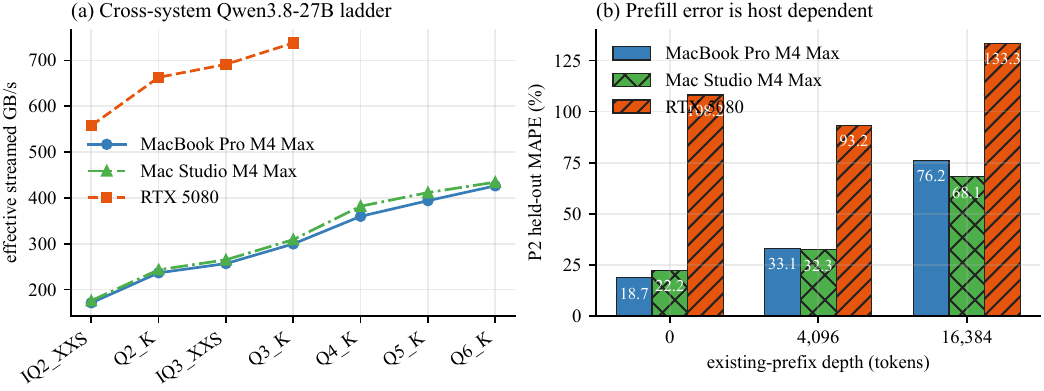}
\caption{Host-separated diagnostics. Left: effective streamed bandwidth for the scored Qwen3.8-27B quantizations at zero prefix; host coverage differs, and IQ2 has 64 layers/26.90B parameters versus 65 layers/27.32B for the other shown files, so the set is a family comparison rather than a strictly identical-network quantization ladder. Right: P2 held-out error when depth-0 coefficients are applied at each existing-prefix depth. Host results are not pooled.}
\end{figure}
\begin{landscape}
\section{Selected model metadata}
Tables~\ref{tab:model-id} and~\ref{tab:model-geometry} jointly reproduce every frozen metadata field used for the 23 unique selected files (22 scored and 1 probe-only). Scored host coverage is explicit because the measurement cohorts are not identical.
\begingroup
\small
\setlength{\tabcolsep}{3.5pt}
\renewcommand{\arraystretch}{1.08}
\begin{longtable}{@{}l l L{0.75in} L{2.7in} L{2.45in} L{0.9in} l r@{}}
\caption{Selected model identity, source, and file size. Sizes are exact bytes; host codes are MB (MacBook Pro M4 Max), MS (Mac Studio M4 Max), RTX (RTX 5080), and ``probe only'' denotes a file excluded from every prediction fit and score.}\label{tab:model-id}\\
\toprule
ID & Split & Scored host(s) & GGUF filename & Repository & Architecture & Quant. & Bytes \\
\midrule
\endfirsthead
\caption[]{Selected model identity, source, and file size. Sizes are exact bytes; host codes are MB (MacBook Pro M4 Max), MS (Mac Studio M4 Max), RTX (RTX 5080), and ``probe only'' denotes a file excluded from every prediction fit and score. (continued)}\\
\toprule
ID & Split & Scored host(s) & GGUF filename & Repository & Architecture & Quant. & Bytes \\
\midrule
\endhead
\midrule
\multicolumn{8}{r}{Continued on next page}\\
\endfoot
\bottomrule
\endlastfoot
M01 & train & MB, MS, RTX & \nolinkurl{gemma-4-12b-it-qat-q4_0.gguf} & \nolinkurl{google/gemma-4-12B-it-qat-q4_0-gguf} & gemma4 & Q4\_0 & 6,975,879,296 \\
M02 & train & MB, MS & \nolinkurl{gemma-4-26B-A4B-it-UD-Q4_K_M.gguf} & \nolinkurl{unsloth/gemma-4-26B-A4B-it-GGUF} & gemma4 & Q4\_K\_M & 16,947,541,728 \\
M03 & test & MS & \nolinkurl{gpt-oss-120b-MXFP4.gguf} & \nolinkurl{ggml-org/gpt-oss-120b-GGUF} & gpt-oss & MXFP4 & 63,387,346,208 \\
M04 & train & MB, MS, RTX & \nolinkurl{gpt-oss-20b-MXFP4.gguf} & \nolinkurl{ggml-org/gpt-oss-20b-GGUF} & gpt-oss & MXFP4 & 12,109,566,624 \\
M05 & test & MB, MS, RTX & \nolinkurl{inclusionAI_Ling-mini-2.0-Q4_K_M.gguf} & \nolinkurl{bartowski/inclusionAI_Ling-mini-2.0-GGUF} & bailingmoe2 & Q4\_K\_M & 9,941,894,336 \\
M06 & test & MB, MS & \nolinkurl{Muse-Glimmer-30B-UD-Q4_K_XL.gguf} & \nolinkurl{unsloth/Muse-Glimmer-30B-GGUF} & muse-glimmer & Q4\_K & 15,878,222,368 \\
M07 & test & MB, MS & \nolinkurl{Nemotron-3-Nano-30B-A3B-Q4_K_M.gguf} & \nolinkurl{unsloth/Nemotron-3-Nano-30B-A3B-GGUF} & nemotron\_h\_moe & Q4\_K\_M & 24,574,373,664 \\
M08 & test & MB, MS, RTX & \nolinkurl{NVIDIA-Nemotron-3-Nano-4B-Q4_K_M.gguf} & \nolinkurl{lmstudio-community/NVIDIA-Nemotron-3-Nano-4B-GGUF} & nemotron\_h & Q4\_K\_M & 2,837,072,896 \\
M09 & train & MB, MS, RTX & \nolinkurl{Qwen3.5-4B-Q4_K_M.gguf} & \nolinkurl{unsloth/Qwen3.5-4B-GGUF} & qwen35 & Q4\_K\_M & 2,740,937,888 \\
M10 & train & RTX & \nolinkurl{Qwen3.5-4B-Q8_0.gguf} & \nolinkurl{unsloth/Qwen3.5-4B-GGUF} & qwen35 & Q8\_0 & 4,482,403,488 \\
M11 & train & MB, MS, RTX & \nolinkurl{Qwen3.5-9B-Q4_K_M.gguf} & \nolinkurl{unsloth/Qwen3.5-9B-GGUF} & qwen35 & Q4\_K\_M & 5,680,522,464 \\
M12 & train & RTX & \nolinkurl{Qwen3.5-9B-Q8_0.gguf} & \nolinkurl{unsloth/Qwen3.5-9B-GGUF} & qwen35 & Q8\_0 & 9,527,502,048 \\
M13 & train & MB, MS & \nolinkurl{Qwen3.6-35B-A3B-UD-Q4_K_M.gguf} & \nolinkurl{unsloth/Qwen3.6-35B-A3B-GGUF} & qwen35moe & Q4\_K\_M & 22,134,528,992 \\
M14 & probe only & -- & \nolinkurl{Qwen3.8-27B-Q8_0.gguf} & \nolinkurl{unsloth/Qwen3.8-27B-GGUF} & qwen35 & Q8\_0 & 29,047,086,048 \\
M15 & train & MB, MS, RTX & \nolinkurl{Qwen3.8-27B-UD-IQ2_XXS.gguf} & \nolinkurl{unsloth/Qwen3.8-27B-GGUF} & qwen35 & IQ2\_XXS & 7,266,070,528 \\
M16 & train & MB, MS, RTX & \nolinkurl{Qwen3.8-27B-UD-IQ3_XXS.gguf} & \nolinkurl{unsloth/Qwen3.8-27B-GGUF} & qwen35 & IQ3\_XXS & 10,934,860,704 \\
M17 & train & MB, MS, RTX & \nolinkurl{Qwen3.8-27B-UD-Q2_K_XL.gguf} & \nolinkurl{unsloth/Qwen3.8-27B-GGUF} & qwen35 & Q2\_K & 9,828,981,664 \\
M18 & train & MB, MS, RTX & \nolinkurl{Qwen3.8-27B-UD-Q3_K_XL.gguf} & \nolinkurl{unsloth/Qwen3.8-27B-GGUF} & qwen35 & Q3\_K & 13,146,393,504 \\
M19 & train & MB, MS & \nolinkurl{Qwen3.8-27B-UD-Q4_K_XL.gguf} & \nolinkurl{unsloth/Qwen3.8-27B-GGUF} & qwen35 & Q4\_K & 17,559,178,144 \\
M20 & train & MB, MS & \nolinkurl{Qwen3.8-27B-UD-Q5_K_XL.gguf} & \nolinkurl{unsloth/Qwen3.8-27B-GGUF} & qwen35 & Q5\_K & 20,876,938,144 \\
M21 & train & MB, MS & \nolinkurl{Qwen3.8-27B-UD-Q6_K_XL.gguf} & \nolinkurl{unsloth/Qwen3.8-27B-GGUF} & qwen35 & Q6\_K & 25,299,061,664 \\
M22 & train & MB, MS, RTX & \nolinkurl{SmolLM3-Q4_K_M.gguf} & \nolinkurl{ggml-org/SmolLM3-3B-GGUF} & smollm3 & Q4\_K\_M & 1,915,305,312 \\
M23 & train & MB, MS, RTX & \nolinkurl{SmolLM3-Q8_0.gguf} & \nolinkurl{ggml-org/SmolLM3-3B-GGUF} & smollm3 & Q8\_0 & 3,275,574,624 \\
\end{longtable}
\endgroup
\end{landscape}
\begin{landscape}
\begingroup
\small
\setlength{\tabcolsep}{3.5pt}
\renewcommand{\arraystretch}{1.08}
\begin{longtable}{@{}l r r r r r r r r r r r r@{}}
\caption{Selected model geometry. KV columns are exact bytes at the measured depths; $k/E$ is routed/total experts.}\label{tab:model-geometry}\\
\toprule
ID & Total P & Active P & Active \% & Layers & Width & Vocab. & KV heads & Head dim. & Used/all & KV@0 & KV@4096 & KV@16384 \\
\midrule
\endfirsthead
\caption[]{Selected model geometry. KV columns are exact bytes at the measured depths; $k/E$ is routed/total experts. (continued)}\\
\toprule
ID & Total P & Active P & Active \% & Layers & Width & Vocab. & KV heads & Head dim. & Used/all & KV@0 & KV@4096 & KV@16384 \\
\midrule
\endhead
\midrule
\multicolumn{13}{r}{Continued on next page}\\
\endfoot
\bottomrule
\endlastfoot
M01 & 11,907,350,576 & 11,907,350,576 & 100.0 & 48 & 3,840 & 262,144 & 8 & 512 & -- & 0 & 738,197,504 & 939,524,096 \\
M02 & 25,233,142,046 & 3,822,527,246 & 15.1 & 30 & 2,816 & 262,144 & 8 & 512 & 8/128 & 0 & 503,316,480 & 754,974,720 \\
M03 & 116,829,156,672 & 5,711,982,912 & 4.9 & 36 & 2,880 & 201,088 & 8 & 64 & 4/128 & 0 & 301,989,888 & 1,207,959,552 \\
M04 & 20,914,757,184 & 4,187,440,704 & 20.0 & 24 & 2,880 & 201,088 & 8 & 64 & 4/32 & 0 & 201,326,592 & 805,306,368 \\
M05 & 16,255,643,392 & 1,432,973,056 & 8.8 & 20 & 2,048 & 157,184 & 4 & 128 & 8/256 & 0 & 167,772,160 & 671,088,640 \\
M06 & 27,854,794,240 & 27,854,794,240 & 100.0 & 52 & 6,656 & 202,048 & 2 & 128 & -- & 0 & 218,103,808 & 872,415,232 \\
M07 & 31,577,940,288 & 3,580,076,352 & 11.3 & 52 & 2,688 & 131,072 & 2 & 128 & 6/128 & 397,934,592 & 423,100,416 & 498,597,888 \\
M08 & 3,973,556,832 & 3,973,556,832 & 100.0 & 42 & 3,136 & 131,072 & 8 & 128 & -- & 616,366,080 & 683,474,944 & 884,801,536 \\
M09 & 4,205,751,296 & 4,205,751,296 & 100.0 & 32 & 2,560 & 248,320 & 4 & 256 & -- & 0 & 536,870,912 & 2,147,483,648 \\
M10 & 4,205,751,296 & 4,205,751,296 & 100.0 & 32 & 2,560 & 248,320 & 4 & 256 & -- & 0 & 536,870,912 & 2,147,483,648 \\
M11 & 8,953,803,264 & 8,953,803,264 & 100.0 & 32 & 4,096 & 248,320 & 4 & 256 & -- & 0 & 536,870,912 & 2,147,483,648 \\
M12 & 8,953,803,264 & 8,953,803,264 & 100.0 & 32 & 4,096 & 248,320 & 4 & 256 & -- & 0 & 536,870,912 & 2,147,483,648 \\
M13 & 34,660,610,688 & 3,454,988,928 & 10.0 & 40 & 2,048 & 248,320 & 2 & 256 & 8/256 & 0 & 335,544,320 & 1,342,177,280 \\
M14 & 27,320,697,856 & 27,320,697,856 & 100.0 & 65 & 5,120 & 248,320 & 4 & 256 & -- & 0 & 1,090,519,040 & 4,362,076,160 \\
M15 & 26,895,998,464 & 26,895,998,464 & 100.0 & 64 & 5,120 & 248,320 & 4 & 256 & -- & 0 & 1,073,741,824 & 4,294,967,296 \\
M16 & 27,320,697,856 & 27,320,697,856 & 100.0 & 65 & 5,120 & 248,320 & 4 & 256 & -- & 0 & 1,090,519,040 & 4,362,076,160 \\
M17 & 27,320,697,856 & 27,320,697,856 & 100.0 & 65 & 5,120 & 248,320 & 4 & 256 & -- & 0 & 1,090,519,040 & 4,362,076,160 \\
M18 & 27,320,697,856 & 27,320,697,856 & 100.0 & 65 & 5,120 & 248,320 & 4 & 256 & -- & 0 & 1,090,519,040 & 4,362,076,160 \\
M19 & 27,320,697,856 & 27,320,697,856 & 100.0 & 65 & 5,120 & 248,320 & 4 & 256 & -- & 0 & 1,090,519,040 & 4,362,076,160 \\
M20 & 27,320,697,856 & 27,320,697,856 & 100.0 & 65 & 5,120 & 248,320 & 4 & 256 & -- & 0 & 1,090,519,040 & 4,362,076,160 \\
M21 & 27,320,697,856 & 27,320,697,856 & 100.0 & 65 & 5,120 & 248,320 & 4 & 256 & -- & 0 & 1,090,519,040 & 4,362,076,160 \\
M22 & 3,075,098,624 & 3,075,098,624 & 100.0 & 36 & 2,048 & 128,256 & 4 & 128 & -- & 0 & 301,989,888 & 1,207,959,552 \\
M23 & 3,075,098,624 & 3,075,098,624 & 100.0 & 36 & 2,048 & 128,256 & 4 & 128 & -- & 0 & 301,989,888 & 1,207,959,552 \\
\end{longtable}
\endgroup
\end{landscape}
\begin{landscape}
\section{All scored decode observations}
These are the 159 rows exported by \nolinkurl{results/predictions.csv}. Throughput is tokens/s; APE is absolute percentage error. Rows are ordered by host, split, model ID, and depth.
\begingroup
\small
\setlength{\tabcolsep}{3.5pt}
\renewcommand{\arraystretch}{1.08}
\begin{longtable}{@{}l l l r r r r r r r@{}}
\caption{Protocol-complete decode measurements and B2 predictions. The retry gate uses the worst decode CV in each host--file invocation. Pre-load is the macOS one-minute POSIX load average or, on Windows, busy-logical-CPU equivalents; it is meaningful within a host but not comparable across hosts.}\label{tab:decode-rows}\\
\toprule
Host & ID & Split & Depth & Measured & B2 & APE \% & Row CV \% & Load & Attempts \\
\midrule
\endfirsthead
\caption[]{Protocol-complete decode measurements and B2 predictions. The retry gate uses the worst decode CV in each host--file invocation. Pre-load is the macOS one-minute POSIX load average or, on Windows, busy-logical-CPU equivalents; it is meaningful within a host but not comparable across hosts. (continued)}\\
\toprule
Host & ID & Split & Depth & Measured & B2 & APE \% & Row CV \% & Load & Attempts \\
\midrule
\endhead
\midrule
\multicolumn{10}{r}{Continued on next page}\\
\endfoot
\bottomrule
\endlastfoot
MacBook Pro M4 Max & M01 & train & 0 & 54.28 & 54.28 & 0.00 & 0.42 & 15.79 & 1 \\
MacBook Pro M4 Max & M01 & train & 4,096 & 49.91 & 49.08 & 1.65 & 0.55 & 15.79 & 1 \\
MacBook Pro M4 Max & M01 & train & 16,384 & 45.34 & 47.84 & 5.50 & 1.54 & 15.79 & 1 \\
MacBook Pro M4 Max & M02 & train & 0 & 89.15 & 124.01 & 39.11 & 0.15 & 16.58 & 1 \\
MacBook Pro M4 Max & M02 & train & 4,096 & 82.91 & 103.69 & 25.07 & 0.42 & 16.58 & 1 \\
MacBook Pro M4 Max & M02 & train & 16,384 & 75.37 & 95.83 & 27.15 & 0.44 & 16.58 & 1 \\
MacBook Pro M4 Max & M04 & train & 0 & 124.28 & 128.17 & 3.13 & 0.29 & 11.83 & 1 \\
MacBook Pro M4 Max & M04 & train & 4,096 & 118.34 & 118.34 & 0.00 & 0.19 & 11.83 & 1 \\
MacBook Pro M4 Max & M04 & train & 16,384 & 102.83 & 96.21 & 6.43 & 0.67 & 11.83 & 1 \\
MacBook Pro M4 Max & M09 & train & 0 & 101.64 & 116.16 & 14.29 & 0.16 & 8.19 & 1 \\
MacBook Pro M4 Max & M09 & train & 4,096 & 97.13 & 97.13 & 0.00 & 0.28 & 8.19 & 1 \\
MacBook Pro M4 Max & M09 & train & 16,384 & 85.05 & 65.13 & 23.42 & 1.25 & 8.19 & 1 \\
MacBook Pro M4 Max & M11 & train & 0 & 66.64 & 56.05 & 15.90 & 1.00 & 8.30 & 1 \\
MacBook Pro M4 Max & M11 & train & 4,096 & 62.95 & 51.21 & 18.65 & 1.81 & 8.30 & 1 \\
MacBook Pro M4 Max & M11 & train & 16,384 & 56.62 & 40.67 & 28.17 & 2.81 & 8.30 & 1 \\
MacBook Pro M4 Max & M13 & train & 0 & 80.85 & 144.30 & 78.49 & 0.29 & 15.62 & 1 \\
MacBook Pro M4 Max & M13 & train & 4,096 & 78.38 & 125.26 & 59.81 & 0.24 & 15.62 & 1 \\
MacBook Pro M4 Max & M13 & train & 16,384 & 72.10 & 89.72 & 24.44 & 1.12 & 15.62 & 1 \\
MacBook Pro M4 Max & M15 & train & 0 & 23.56 & 23.56 & 0.00 & 1.20 & 16.52 & 1 \\
MacBook Pro M4 Max & M15 & train & 4,096 & 20.41 & 20.53 & 0.59 & 2.70 & 16.52 & 1 \\
MacBook Pro M4 Max & M15 & train & 16,384 & 19.71 & 14.81 & 24.88 & 1.72 & 16.52 & 1 \\
MacBook Pro M4 Max & M16 & train & 0 & 23.49 & 24.42 & 3.94 & 1.09 & 14.84 & 1 \\
MacBook Pro M4 Max & M16 & train & 4,096 & 22.20 & 22.20 & 0.00 & 1.59 & 14.84 & 1 \\
MacBook Pro M4 Max & M16 & train & 16,384 & 19.63 & 17.45 & 11.07 & 1.65 & 14.84 & 1 \\
MacBook Pro M4 Max & M17 & train & 0 & 24.08 & 25.71 & 6.75 & 1.90 & 11.04 & 1 \\
MacBook Pro M4 Max & M17 & train & 4,096 & 23.14 & 23.14 & 0.00 & 2.39 & 11.04 & 1 \\
MacBook Pro M4 Max & M17 & train & 16,384 & 21.30 & 17.81 & 16.41 & 1.65 & 11.04 & 1 \\
MacBook Pro M4 Max & M18 & train & 0 & 22.80 & 23.29 & 2.12 & 2.60 & 11.42 & 3 \\
MacBook Pro M4 Max & M18 & train & 4,096 & 21.89 & 21.50 & 1.75 & 0.91 & 11.42 & 3 \\
MacBook Pro M4 Max & M18 & train & 16,384 & 17.49 & 17.49 & 0.00 & 2.94 & 11.42 & 3 \\
MacBook Pro M4 Max & M19 & train & 0 & 20.50 & 20.92 & 2.05 & 2.11 & 13.06 & 3 \\
MacBook Pro M4 Max & M19 & train & 4,096 & 19.70 & 19.70 & 0.00 & 1.97 & 13.06 & 3 \\
MacBook Pro M4 Max & M19 & train & 16,384 & 16.77 & 16.76 & 0.10 & 1.47 & 13.06 & 3 \\
MacBook Pro M4 Max & M20 & train & 0 & 18.88 & 19.13 & 1.31 & 1.57 & 17.66 & 1 \\
MacBook Pro M4 Max & M20 & train & 4,096 & 18.33 & 18.18 & 0.83 & 2.21 & 17.66 & 1 \\
MacBook Pro M4 Max & M20 & train & 16,384 & 15.82 & 15.82 & 0.00 & 2.67 & 17.66 & 1 \\
MacBook Pro M4 Max & M21 & train & 0 & 16.86 & 16.86 & 0.00 & 0.37 & 17.18 & 1 \\
MacBook Pro M4 Max & M21 & train & 4,096 & 15.92 & 16.16 & 1.50 & 1.42 & 17.18 & 1 \\
MacBook Pro M4 Max & M21 & train & 16,384 & 14.97 & 14.38 & 3.96 & 0.92 & 17.18 & 1 \\
MacBook Pro M4 Max & M22 & train & 0 & 169.50 & 166.23 & 1.93 & 0.46 & 13.91 & 1 \\
MacBook Pro M4 Max & M22 & train & 4,096 & 149.47 & 143.59 & 3.93 & 0.31 & 13.91 & 1 \\
MacBook Pro M4 Max & M22 & train & 16,384 & 115.22 & 101.94 & 11.53 & 0.40 & 13.91 & 1 \\
MacBook Pro M4 Max & M23 & train & 0 & 118.87 & 119.53 & 0.55 & 0.40 & 13.92 & 1 \\
MacBook Pro M4 Max & M23 & train & 4,096 & 109.44 & 109.44 & 0.00 & 0.41 & 13.92 & 1 \\
MacBook Pro M4 Max & M23 & train & 16,384 & 88.22 & 87.32 & 1.01 & 1.00 & 13.92 & 1 \\
MacBook Pro M4 Max & M05 & test & 0 & 242.33 & 363.29 & 49.92 & 1.24 & 11.64 & 1 \\
MacBook Pro M4 Max & M05 & test & 4,096 & 219.67 & 304.92 & 38.81 & 0.84 & 11.64 & 1 \\
MacBook Pro M4 Max & M05 & test & 16,384 & 175.60 & 205.75 & 17.17 & 0.37 & 11.64 & 1 \\
MacBook Pro M4 Max & M06 & test & 0 & 26.23 & 23.14 & 11.81 & 1.92 & 8.93 & 3 \\
MacBook Pro M4 Max & M06 & test & 4,096 & 23.30 & 22.82 & 2.05 & 2.13 & 8.93 & 3 \\
MacBook Pro M4 Max & M06 & test & 16,384 & 22.55 & 21.93 & 2.74 & 3.03 & 8.93 & 3 \\
MacBook Pro M4 Max & M07 & test & 0 & 92.12 & 100.00 & 8.55 & 0.30 & 10.01 & 1 \\
MacBook Pro M4 Max & M07 & test & 4,096 & 90.68 & 99.21 & 9.41 & 0.49 & 10.01 & 1 \\
MacBook Pro M4 Max & M07 & test & 16,384 & 85.16 & 96.93 & 13.83 & 0.68 & 10.01 & 1 \\
MacBook Pro M4 Max & M08 & test & 0 & 93.29 & 92.19 & 1.18 & 0.30 & 8.78 & 1 \\
MacBook Pro M4 Max & M08 & test & 4,096 & 91.41 & 90.44 & 1.07 & 0.10 & 8.78 & 1 \\
MacBook Pro M4 Max & M08 & test & 16,384 & 84.86 & 85.55 & 0.80 & 0.42 & 8.78 & 1 \\
Mac Studio M4 Max & M01 & train & 0 & 54.46 & 54.46 & 0.00 & 0.41 & 10.38 & 1 \\
Mac Studio M4 Max & M01 & train & 4,096 & 50.62 & 49.25 & 2.71 & 2.06 & 10.38 & 1 \\
Mac Studio M4 Max & M01 & train & 16,384 & 47.37 & 47.99 & 1.31 & 0.16 & 10.38 & 1 \\
Mac Studio M4 Max & M02 & train & 0 & 88.10 & 124.32 & 41.11 & 0.22 & 4.69 & 1 \\
Mac Studio M4 Max & M02 & train & 4,096 & 79.77 & 103.94 & 30.31 & 2.30 & 4.69 & 1 \\
Mac Studio M4 Max & M02 & train & 16,384 & 75.49 & 96.07 & 27.26 & 0.24 & 4.69 & 1 \\
Mac Studio M4 Max & M04 & train & 0 & 123.10 & 126.93 & 3.11 & 0.36 & 3.28 & 1 \\
Mac Studio M4 Max & M04 & train & 4,096 & 117.20 & 117.20 & 0.00 & 0.81 & 3.28 & 1 \\
Mac Studio M4 Max & M04 & train & 16,384 & 105.36 & 95.28 & 9.57 & 0.31 & 3.28 & 1 \\
Mac Studio M4 Max & M09 & train & 0 & 101.41 & 116.45 & 14.83 & 0.42 & 4.06 & 1 \\
Mac Studio M4 Max & M09 & train & 4,096 & 97.37 & 97.37 & 0.00 & 1.90 & 4.06 & 1 \\
Mac Studio M4 Max & M09 & train & 16,384 & 88.73 & 65.29 & 26.41 & 0.23 & 4.06 & 1 \\
Mac Studio M4 Max & M11 & train & 0 & 68.03 & 56.19 & 17.41 & 0.12 & 3.12 & 1 \\
Mac Studio M4 Max & M11 & train & 4,096 & 66.41 & 51.34 & 22.69 & 0.14 & 3.12 & 1 \\
Mac Studio M4 Max & M11 & train & 16,384 & 61.83 & 40.77 & 34.06 & 0.18 & 3.12 & 1 \\
Mac Studio M4 Max & M13 & train & 0 & 81.12 & 144.66 & 78.32 & 0.68 & 3.80 & 1 \\
Mac Studio M4 Max & M13 & train & 4,096 & 77.79 & 125.56 & 61.42 & 1.17 & 3.80 & 1 \\
Mac Studio M4 Max & M13 & train & 16,384 & 73.18 & 89.94 & 22.91 & 0.87 & 3.80 & 1 \\
Mac Studio M4 Max & M15 & train & 0 & 24.22 & 27.31 & 12.74 & 0.37 & 2.98 & 1 \\
Mac Studio M4 Max & M15 & train & 4,096 & 23.79 & 23.79 & 0.00 & 0.95 & 2.98 & 1 \\
Mac Studio M4 Max & M15 & train & 16,384 & 22.40 & 17.16 & 23.36 & 0.30 & 2.98 & 1 \\
Mac Studio M4 Max & M16 & train & 0 & 24.24 & 26.19 & 8.02 & 0.31 & 2.14 & 1 \\
Mac Studio M4 Max & M16 & train & 4,096 & 23.81 & 23.81 & 0.00 & 0.69 & 2.14 & 1 \\
Mac Studio M4 Max & M16 & train & 16,384 & 22.34 & 18.72 & 16.18 & 0.55 & 2.14 & 1 \\
Mac Studio M4 Max & M17 & train & 0 & 24.76 & 26.91 & 8.68 & 0.45 & 2.92 & 1 \\
Mac Studio M4 Max & M17 & train & 4,096 & 24.23 & 24.23 & 0.00 & 0.43 & 2.92 & 1 \\
Mac Studio M4 Max & M17 & train & 16,384 & 22.83 & 18.64 & 18.35 & 0.71 & 2.92 & 1 \\
Mac Studio M4 Max & M18 & train & 0 & 23.52 & 24.99 & 6.23 & 0.68 & 4.39 & 1 \\
Mac Studio M4 Max & M18 & train & 4,096 & 23.07 & 23.07 & 0.00 & 0.47 & 4.39 & 1 \\
Mac Studio M4 Max & M18 & train & 16,384 & 21.77 & 18.76 & 13.81 & 0.66 & 4.39 & 1 \\
Mac Studio M4 Max & M19 & train & 0 & 21.76 & 22.86 & 5.07 & 1.00 & 6.44 & 1 \\
Mac Studio M4 Max & M19 & train & 4,096 & 21.52 & 21.52 & 0.00 & 0.80 & 6.44 & 1 \\
Mac Studio M4 Max & M19 & train & 16,384 & 20.41 & 18.31 & 10.29 & 1.01 & 6.44 & 1 \\
Mac Studio M4 Max & M20 & train & 0 & 19.72 & 20.46 & 3.76 & 0.13 & 2.24 & 1 \\
Mac Studio M4 Max & M20 & train & 4,096 & 19.45 & 19.45 & 0.00 & 0.33 & 2.24 & 1 \\
Mac Studio M4 Max & M20 & train & 16,384 & 18.54 & 16.93 & 8.71 & 0.14 & 2.24 & 1 \\
Mac Studio M4 Max & M21 & train & 0 & 17.18 & 17.67 & 2.87 & 0.38 & 2.99 & 1 \\
Mac Studio M4 Max & M21 & train & 4,096 & 16.94 & 16.94 & 0.00 & 0.30 & 2.99 & 1 \\
Mac Studio M4 Max & M21 & train & 16,384 & 16.29 & 15.07 & 7.49 & 0.17 & 2.99 & 1 \\
Mac Studio M4 Max & M22 & train & 0 & 168.55 & 166.64 & 1.13 & 0.42 & 4.33 & 2 \\
Mac Studio M4 Max & M22 & train & 4,096 & 149.85 & 143.95 & 3.94 & 1.20 & 4.33 & 2 \\
Mac Studio M4 Max & M22 & train & 16,384 & 115.65 & 102.19 & 11.64 & 0.24 & 4.33 & 2 \\
Mac Studio M4 Max & M23 & train & 0 & 119.69 & 120.58 & 0.75 & 0.25 & 5.97 & 1 \\
Mac Studio M4 Max & M23 & train & 4,096 & 110.40 & 110.40 & 0.00 & 0.08 & 5.97 & 1 \\
Mac Studio M4 Max & M23 & train & 16,384 & 89.60 & 88.10 & 1.68 & 2.06 & 5.97 & 1 \\
Mac Studio M4 Max & M03 & test & 0 & 85.42 & 99.30 & 16.25 & 0.71 & 5.93 & 1 \\
Mac Studio M4 Max & M03 & test & 4,096 & 81.13 & 90.48 & 11.54 & 1.09 & 5.93 & 1 \\
Mac Studio M4 Max & M03 & test & 16,384 & 72.80 & 71.45 & 1.85 & 0.85 & 5.93 & 1 \\
Mac Studio M4 Max & M05 & test & 0 & 238.62 & 364.19 & 52.62 & 1.31 & 3.14 & 1 \\
Mac Studio M4 Max & M05 & test & 4,096 & 213.24 & 305.67 & 43.34 & 1.52 & 3.14 & 1 \\
Mac Studio M4 Max & M05 & test & 16,384 & 172.28 & 206.25 & 19.72 & 0.74 & 3.14 & 1 \\
Mac Studio M4 Max & M06 & test & 0 & 28.50 & 25.28 & 11.30 & 0.26 & 5.50 & 1 \\
Mac Studio M4 Max & M06 & test & 4,096 & 27.82 & 24.94 & 10.35 & 0.58 & 5.50 & 1 \\
Mac Studio M4 Max & M06 & test & 16,384 & 27.07 & 23.96 & 11.49 & 0.27 & 5.50 & 1 \\
Mac Studio M4 Max & M07 & test & 0 & 91.77 & 100.24 & 9.24 & 0.66 & 2.65 & 1 \\
Mac Studio M4 Max & M07 & test & 4,096 & 90.27 & 99.46 & 10.18 & 0.69 & 2.65 & 1 \\
Mac Studio M4 Max & M07 & test & 16,384 & 86.07 & 97.17 & 12.90 & 0.47 & 2.65 & 1 \\
Mac Studio M4 Max & M08 & test & 0 & 93.58 & 92.42 & 1.24 & 0.25 & 4.15 & 1 \\
Mac Studio M4 Max & M08 & test & 4,096 & 91.56 & 90.66 & 0.99 & 0.82 & 4.15 & 1 \\
Mac Studio M4 Max & M08 & test & 16,384 & 87.96 & 85.76 & 2.50 & 0.40 & 4.15 & 1 \\
RTX 5080 & M01 & train & 0 & 98.27 & 104.03 & 5.86 & 0.31 & 3.58 & 1 \\
RTX 5080 & M01 & train & 4,096 & 94.08 & 94.08 & 0.00 & 0.64 & 3.58 & 1 \\
RTX 5080 & M01 & train & 16,384 & 92.53 & 91.68 & 0.91 & 0.45 & 3.58 & 1 \\
RTX 5080 & M04 & train & 0 & 229.65 & 236.51 & 2.99 & 0.48 & 8.58 & 2 \\
RTX 5080 & M04 & train & 4,096 & 218.37 & 218.37 & 0.00 & 0.54 & 8.58 & 2 \\
RTX 5080 & M04 & train & 16,384 & 203.03 & 177.54 & 12.56 & 1.58 & 8.58 & 2 \\
RTX 5080 & M09 & train & 0 & 202.01 & 241.17 & 19.39 & 0.33 & 1.02 & 1 \\
RTX 5080 & M09 & train & 4,096 & 198.97 & 201.67 & 1.36 & 0.50 & 1.02 & 1 \\
RTX 5080 & M09 & train & 16,384 & 181.22 & 135.22 & 25.38 & 0.52 & 1.02 & 1 \\
RTX 5080 & M10 & train & 0 & 147.62 & 163.32 & 10.64 & 0.15 & 1.60 & 1 \\
RTX 5080 & M10 & train & 4,096 & 145.85 & 145.85 & 0.00 & 0.50 & 1.60 & 1 \\
RTX 5080 & M10 & train & 16,384 & 136.04 & 110.42 & 18.83 & 0.34 & 1.60 & 1 \\
RTX 5080 & M11 & train & 0 & 132.95 & 116.37 & 12.47 & 0.15 & 1.66 & 1 \\
RTX 5080 & M11 & train & 4,096 & 131.63 & 106.32 & 19.23 & 0.43 & 1.66 & 1 \\
RTX 5080 & M11 & train & 16,384 & 123.69 & 84.44 & 31.73 & 0.43 & 1.66 & 1 \\
RTX 5080 & M12 & train & 0 & 88.98 & 76.84 & 13.65 & 0.09 & 1.47 & 1 \\
RTX 5080 & M12 & train & 4,096 & 88.23 & 72.74 & 17.55 & 0.23 & 1.47 & 1 \\
RTX 5080 & M12 & train & 16,384 & 84.85 & 62.70 & 26.10 & 0.24 & 1.47 & 1 \\
RTX 5080 & M15 & train & 0 & 76.76 & 86.74 & 13.01 & 0.33 & 11.26 & 2 \\
RTX 5080 & M15 & train & 4,096 & 75.58 & 75.58 & 0.00 & 0.49 & 11.26 & 2 \\
RTX 5080 & M15 & train & 16,384 & 70.36 & 54.52 & 22.51 & 1.39 & 11.26 & 2 \\
RTX 5080 & M16 & train & 0 & 63.24 & 68.69 & 8.62 & 0.28 & 14.78 & 1 \\
RTX 5080 & M16 & train & 4,096 & 62.46 & 62.46 & 0.00 & 0.41 & 14.78 & 1 \\
RTX 5080 & M16 & train & 16,384 & 58.63 & 49.10 & 16.25 & 1.50 & 14.78 & 1 \\
RTX 5080 & M17 & train & 0 & 67.48 & 73.66 & 9.15 & 0.30 & 12.10 & 1 \\
RTX 5080 & M17 & train & 4,096 & 66.30 & 66.30 & 0.00 & 0.41 & 12.10 & 1 \\
RTX 5080 & M17 & train & 16,384 & 62.22 & 51.02 & 18.00 & 1.65 & 12.10 & 1 \\
RTX 5080 & M18 & train & 0 & 56.15 & 60.10 & 7.02 & 0.22 & 7.74 & 1 \\
RTX 5080 & M18 & train & 4,096 & 55.49 & 55.49 & 0.00 & 0.36 & 7.74 & 1 \\
RTX 5080 & M18 & train & 16,384 & 52.65 & 45.12 & 14.30 & 1.02 & 7.74 & 1 \\
RTX 5080 & M22 & train & 0 & 306.16 & 345.13 & 12.73 & 0.49 & 1.41 & 1 \\
RTX 5080 & M22 & train & 4,096 & 272.17 & 298.13 & 9.54 & 0.74 & 1.41 & 1 \\
RTX 5080 & M22 & train & 16,384 & 211.65 & 211.65 & 0.00 & 0.48 & 1.41 & 1 \\
RTX 5080 & M23 & train & 0 & 209.46 & 223.49 & 6.70 & 0.19 & 0.70 & 1 \\
RTX 5080 & M23 & train & 4,096 & 192.60 & 204.63 & 6.24 & 0.33 & 0.70 & 1 \\
RTX 5080 & M23 & train & 16,384 & 160.88 & 163.28 & 1.49 & 0.29 & 0.70 & 1 \\
RTX 5080 & M05 & test & 0 & 446.42 & 754.26 & 68.96 & 0.76 & 2.05 & 1 \\
RTX 5080 & M05 & test & 4,096 & 406.19 & 633.07 & 55.85 & 1.35 & 2.05 & 1 \\
RTX 5080 & M05 & test & 16,384 & 326.38 & 427.17 & 30.88 & 1.04 & 2.05 & 1 \\
RTX 5080 & M08 & test & 0 & 239.52 & 191.41 & 20.09 & 0.41 & 1.86 & 2 \\
RTX 5080 & M08 & test & 4,096 & 236.79 & 187.76 & 20.70 & 0.29 & 1.86 & 2 \\
RTX 5080 & M08 & test & 16,384 & 223.15 & 177.61 & 20.41 & 1.01 & 1.86 & 2 \\
\end{longtable}
\endgroup
\end{landscape}
\begin{landscape}
\section{All scored prefill observations}
These are the 159 rows exported by \nolinkurl{results/predictions_prefill.csv}. P1 has one host coefficient; P2 has one host-by-quantization coefficient. Both are fit only at depth 0. The row CV column is reported rather than used as an inclusion filter.
\begingroup
\small
\setlength{\tabcolsep}{3.5pt}
\renewcommand{\arraystretch}{1.08}
\begin{longtable}{@{}l l l r l r r r r r r@{}}
\caption{Protocol-complete prefill measurements and predictions. The retry decision was decode-gated, not prefill-gated; depth-0 rows are the primary evaluation and larger depths are scope diagnostics.}\label{tab:prefill-rows}\\
\toprule
Host & ID & Split & Depth & Scope & Measured & P1 & P1 APE & P2 & P2 APE & Row CV \% \\
\midrule
\endfirsthead
\caption[]{Protocol-complete prefill measurements and predictions. The retry decision was decode-gated, not prefill-gated; depth-0 rows are the primary evaluation and larger depths are scope diagnostics. (continued)}\\
\toprule
Host & ID & Split & Depth & Scope & Measured & P1 & P1 APE & P2 & P2 APE & Row CV \% \\
\midrule
\endhead
\midrule
\multicolumn{11}{r}{Continued on next page}\\
\endfoot
\bottomrule
\endlastfoot
MacBook Pro M4 Max & M01 & train & 0 & \textbf{primary} & 589.73 & 547.32 & 7.19 & 589.73 & 0.00 & 0.05 \\
MacBook Pro M4 Max & M01 & train & 4,096 & diagnostic & 491.90 & 547.32 & 11.27 & 589.73 & 19.89 & 1.01 \\
MacBook Pro M4 Max & M01 & train & 16,384 & diagnostic & 382.76 & 547.32 & 42.99 & 589.73 & 54.07 & 3.47 \\
MacBook Pro M4 Max & M02 & train & 0 & \textbf{primary} & 1418.29 & 1704.93 & 20.21 & 1610.29 & 13.54 & 0.38 \\
MacBook Pro M4 Max & M02 & train & 4,096 & diagnostic & 1145.69 & 1704.93 & 48.81 & 1610.29 & 40.55 & 2.81 \\
MacBook Pro M4 Max & M02 & train & 16,384 & diagnostic & 741.02 & 1704.93 & 130.08 & 1610.29 & 117.31 & 2.63 \\
MacBook Pro M4 Max & M04 & train & 0 & \textbf{primary} & 1589.84 & 1556.35 & 2.11 & 1589.84 & 0.00 & 0.27 \\
MacBook Pro M4 Max & M04 & train & 4,096 & diagnostic & 1372.89 & 1556.35 & 13.36 & 1589.84 & 15.80 & 2.79 \\
MacBook Pro M4 Max & M04 & train & 16,384 & diagnostic & 848.96 & 1556.35 & 83.33 & 1589.84 & 87.27 & 0.72 \\
MacBook Pro M4 Max & M09 & train & 0 & \textbf{primary} & 1463.56 & 1549.58 & 5.88 & 1463.56 & 0.00 & 0.07 \\
MacBook Pro M4 Max & M09 & train & 4,096 & diagnostic & 1310.45 & 1549.58 & 18.25 & 1463.56 & 11.68 & 0.40 \\
MacBook Pro M4 Max & M09 & train & 16,384 & diagnostic & 941.78 & 1549.58 & 64.54 & 1463.56 & 55.40 & 0.89 \\
MacBook Pro M4 Max & M11 & train & 0 & \textbf{primary} & 819.16 & 727.86 & 11.14 & 687.46 & 16.08 & 0.13 \\
MacBook Pro M4 Max & M11 & train & 4,096 & diagnostic & 727.07 & 727.86 & 0.11 & 687.46 & 5.45 & 0.39 \\
MacBook Pro M4 Max & M11 & train & 16,384 & diagnostic & 564.60 & 727.86 & 28.92 & 687.46 & 21.76 & 1.59 \\
MacBook Pro M4 Max & M13 & train & 0 & \textbf{primary} & 1414.02 & 1886.30 & 33.40 & 1781.59 & 25.99 & 0.67 \\
MacBook Pro M4 Max & M13 & train & 4,096 & diagnostic & 1189.17 & 1886.30 & 58.62 & 1781.59 & 49.82 & 1.38 \\
MacBook Pro M4 Max & M13 & train & 16,384 & diagnostic & 880.48 & 1886.30 & 114.24 & 1781.59 & 102.34 & 1.39 \\
MacBook Pro M4 Max & M15 & train & 0 & \textbf{primary} & 239.52 & 242.31 & 1.17 & 239.52 & 0.00 & 2.73 \\
MacBook Pro M4 Max & M15 & train & 4,096 & diagnostic & 196.04 & 242.31 & 23.60 & 239.52 & 22.18 & 0.31 \\
MacBook Pro M4 Max & M15 & train & 16,384 & diagnostic & 158.60 & 242.31 & 52.78 & 239.52 & 51.02 & 1.33 \\
MacBook Pro M4 Max & M16 & train & 0 & \textbf{primary} & 238.54 & 238.54 & 0.00 & 238.54 & 0.00 & 2.42 \\
MacBook Pro M4 Max & M16 & train & 4,096 & diagnostic & 203.40 & 238.54 & 17.28 & 238.54 & 17.28 & 1.81 \\
MacBook Pro M4 Max & M16 & train & 16,384 & diagnostic & 172.91 & 238.54 & 37.96 & 238.54 & 37.96 & 1.74 \\
MacBook Pro M4 Max & M17 & train & 0 & \textbf{primary} & 235.14 & 238.54 & 1.45 & 235.14 & 0.00 & 1.72 \\
MacBook Pro M4 Max & M17 & train & 4,096 & diagnostic & 207.61 & 238.54 & 14.90 & 235.14 & 13.26 & 2.63 \\
MacBook Pro M4 Max & M17 & train & 16,384 & diagnostic & 180.12 & 238.54 & 32.43 & 235.14 & 30.54 & 1.42 \\
MacBook Pro M4 Max & M18 & train & 0 & \textbf{primary} & 241.51 & 238.54 & 1.23 & 241.51 & 0.00 & 2.59 \\
MacBook Pro M4 Max & M18 & train & 4,096 & diagnostic & 202.71 & 238.54 & 17.67 & 241.51 & 19.14 & 0.93 \\
MacBook Pro M4 Max & M18 & train & 16,384 & diagnostic & 164.64 & 238.54 & 44.89 & 241.51 & 46.69 & 2.76 \\
MacBook Pro M4 Max & M19 & train & 0 & \textbf{primary} & 224.91 & 238.54 & 6.06 & 224.91 & 0.00 & 3.91 \\
MacBook Pro M4 Max & M19 & train & 4,096 & diagnostic & 191.29 & 238.54 & 24.70 & 224.91 & 17.58 & 0.80 \\
MacBook Pro M4 Max & M19 & train & 16,384 & diagnostic & 162.37 & 238.54 & 46.91 & 224.91 & 38.52 & 2.61 \\
MacBook Pro M4 Max & M20 & train & 0 & \textbf{primary} & 225.20 & 238.54 & 5.92 & 225.20 & 0.00 & 2.87 \\
MacBook Pro M4 Max & M20 & train & 4,096 & diagnostic & 196.00 & 238.54 & 21.70 & 225.20 & 14.90 & 2.15 \\
MacBook Pro M4 Max & M20 & train & 16,384 & diagnostic & 174.79 & 238.54 & 36.47 & 225.20 & 28.84 & 1.41 \\
MacBook Pro M4 Max & M21 & train & 0 & \textbf{primary} & 239.03 & 238.54 & 0.20 & 239.03 & 0.00 & 2.54 \\
MacBook Pro M4 Max & M21 & train & 4,096 & diagnostic & 208.07 & 238.54 & 14.64 & 239.03 & 14.88 & 2.22 \\
MacBook Pro M4 Max & M21 & train & 16,384 & diagnostic & 181.38 & 238.54 & 31.52 & 239.03 & 31.78 & 1.35 \\
MacBook Pro M4 Max & M22 & train & 0 & \textbf{primary} & 2135.94 & 2119.33 & 0.78 & 2001.68 & 6.29 & 0.04 \\
MacBook Pro M4 Max & M22 & train & 4,096 & diagnostic & 1653.76 & 2119.33 & 28.15 & 2001.68 & 21.04 & 2.82 \\
MacBook Pro M4 Max & M22 & train & 16,384 & diagnostic & 965.07 & 2119.33 & 119.60 & 2001.68 & 107.41 & 1.34 \\
MacBook Pro M4 Max & M23 & train & 0 & \textbf{primary} & 2239.31 & 2119.33 & 5.36 & 2239.31 & 0.00 & 0.12 \\
MacBook Pro M4 Max & M23 & train & 4,096 & diagnostic & 1735.34 & 2119.33 & 22.13 & 2239.31 & 29.04 & 0.12 \\
MacBook Pro M4 Max & M23 & train & 16,384 & diagnostic & 892.27 & 2119.33 & 137.52 & 2239.31 & 150.97 & 1.34 \\
MacBook Pro M4 Max & M05 & test & 0 & \textbf{primary} & 4208.20 & 4547.98 & 8.07 & 4295.53 & 2.08 & 0.42 \\
MacBook Pro M4 Max & M05 & test & 4,096 & diagnostic & 3300.39 & 4547.98 & 37.80 & 4295.53 & 30.15 & 0.62 \\
MacBook Pro M4 Max & M05 & test & 16,384 & diagnostic & 1848.16 & 4547.98 & 146.08 & 4295.53 & 132.42 & 1.94 \\
MacBook Pro M4 Max & M06 & test & 0 & \textbf{primary} & 236.65 & 233.97 & 1.13 & 220.60 & 6.78 & 3.52 \\
MacBook Pro M4 Max & M06 & test & 4,096 & diagnostic & 187.15 & 233.97 & 25.01 & 220.60 & 17.87 & 3.95 \\
MacBook Pro M4 Max & M06 & test & 16,384 & diagnostic & 176.87 & 233.97 & 32.28 & 220.60 & 24.72 & 1.62 \\
MacBook Pro M4 Max & M07 & test & 0 & \textbf{primary} & 1203.59 & 1820.39 & 51.25 & 1719.34 & 42.85 & 0.45 \\
MacBook Pro M4 Max & M07 & test & 4,096 & diagnostic & 1106.68 & 1820.39 & 64.49 & 1719.34 & 55.36 & 3.64 \\
MacBook Pro M4 Max & M07 & test & 16,384 & diagnostic & 895.52 & 1820.39 & 103.28 & 1719.34 & 91.99 & 1.19 \\
MacBook Pro M4 Max & M08 & test & 0 & \textbf{primary} & 1259.18 & 1640.13 & 30.25 & 1549.08 & 23.02 & 4.84 \\
MacBook Pro M4 Max & M08 & test & 4,096 & diagnostic & 1200.41 & 1640.13 & 36.63 & 1549.08 & 29.05 & 3.30 \\
MacBook Pro M4 Max & M08 & test & 16,384 & diagnostic & 995.75 & 1640.13 & 64.71 & 1549.08 & 55.57 & 1.37 \\
Mac Studio M4 Max & M01 & train & 0 & \textbf{primary} & 588.97 & 565.21 & 4.04 & 588.97 & 0.00 & 0.45 \\
Mac Studio M4 Max & M01 & train & 4,096 & diagnostic & 521.77 & 565.21 & 8.32 & 588.97 & 12.88 & 0.20 \\
Mac Studio M4 Max & M01 & train & 16,384 & diagnostic & 413.59 & 565.21 & 36.66 & 588.97 & 42.41 & 0.91 \\
Mac Studio M4 Max & M02 & train & 0 & \textbf{primary} & 1408.11 & 1760.64 & 25.04 & 1624.90 & 15.40 & 0.75 \\
Mac Studio M4 Max & M02 & train & 4,096 & diagnostic & 1181.84 & 1760.64 & 48.97 & 1624.90 & 37.49 & 0.93 \\
Mac Studio M4 Max & M02 & train & 16,384 & diagnostic & 869.98 & 1760.64 & 102.38 & 1624.90 & 86.77 & 0.36 \\
Mac Studio M4 Max & M04 & train & 0 & \textbf{primary} & 1576.97 & 1607.21 & 1.92 & 1576.97 & 0.00 & 0.79 \\
Mac Studio M4 Max & M04 & train & 4,096 & diagnostic & 1389.95 & 1607.21 & 15.63 & 1576.97 & 13.45 & 0.27 \\
Mac Studio M4 Max & M04 & train & 16,384 & diagnostic & 995.24 & 1607.21 & 61.49 & 1576.97 & 58.45 & 0.28 \\
Mac Studio M4 Max & M09 & train & 0 & \textbf{primary} & 1476.84 & 1600.21 & 8.35 & 1476.84 & 0.00 & 0.12 \\
Mac Studio M4 Max & M09 & train & 4,096 & diagnostic & 1357.24 & 1600.21 & 17.90 & 1476.84 & 8.81 & 0.20 \\
Mac Studio M4 Max & M09 & train & 16,384 & diagnostic & 1083.61 & 1600.21 & 47.67 & 1476.84 & 36.29 & 0.16 \\
Mac Studio M4 Max & M11 & train & 0 & \textbf{primary} & 832.13 & 751.65 & 9.67 & 693.70 & 16.64 & 0.10 \\
Mac Studio M4 Max & M11 & train & 4,096 & diagnostic & 792.01 & 751.65 & 5.10 & 693.70 & 12.41 & 0.12 \\
Mac Studio M4 Max & M11 & train & 16,384 & diagnostic & 691.18 & 751.65 & 8.75 & 693.70 & 0.36 & 0.12 \\
Mac Studio M4 Max & M13 & train & 0 & \textbf{primary} & 1423.47 & 1947.94 & 36.84 & 1797.76 & 26.29 & 0.75 \\
Mac Studio M4 Max & M13 & train & 4,096 & diagnostic & 1279.33 & 1947.94 & 52.26 & 1797.76 & 40.52 & 0.47 \\
Mac Studio M4 Max & M13 & train & 16,384 & diagnostic & 994.74 & 1947.94 & 95.82 & 1797.76 & 80.73 & 0.21 \\
Mac Studio M4 Max & M15 & train & 0 & \textbf{primary} & 251.74 & 250.23 & 0.60 & 251.74 & 0.00 & 0.07 \\
Mac Studio M4 Max & M15 & train & 4,096 & diagnostic & 240.96 & 250.23 & 3.84 & 251.74 & 4.47 & 0.03 \\
Mac Studio M4 Max & M15 & train & 16,384 & diagnostic & 213.03 & 250.23 & 17.46 & 251.74 & 18.17 & 0.07 \\
Mac Studio M4 Max & M16 & train & 0 & \textbf{primary} & 250.39 & 246.34 & 1.62 & 250.39 & 0.00 & 0.03 \\
Mac Studio M4 Max & M16 & train & 4,096 & diagnostic & 239.79 & 246.34 & 2.73 & 250.39 & 4.42 & 0.04 \\
Mac Studio M4 Max & M16 & train & 16,384 & diagnostic & 211.89 & 246.34 & 16.26 & 250.39 & 18.17 & 0.03 \\
Mac Studio M4 Max & M17 & train & 0 & \textbf{primary} & 251.10 & 246.34 & 1.90 & 251.10 & 0.00 & 0.03 \\
Mac Studio M4 Max & M17 & train & 4,096 & diagnostic & 240.25 & 246.34 & 2.53 & 251.10 & 4.52 & 0.06 \\
Mac Studio M4 Max & M17 & train & 16,384 & diagnostic & 212.36 & 246.34 & 16.00 & 251.10 & 18.25 & 0.01 \\
Mac Studio M4 Max & M18 & train & 0 & \textbf{primary} & 249.04 & 246.34 & 1.09 & 249.04 & 0.00 & 0.02 \\
Mac Studio M4 Max & M18 & train & 4,096 & diagnostic & 238.52 & 246.34 & 3.28 & 249.04 & 4.41 & 0.04 \\
Mac Studio M4 Max & M18 & train & 16,384 & diagnostic & 211.04 & 246.34 & 16.73 & 249.04 & 18.01 & 0.02 \\
Mac Studio M4 Max & M19 & train & 0 & \textbf{primary} & 236.56 & 246.34 & 4.13 & 236.56 & 0.00 & 0.07 \\
Mac Studio M4 Max & M19 & train & 4,096 & diagnostic & 227.01 & 246.34 & 8.51 & 236.56 & 4.21 & 0.06 \\
Mac Studio M4 Max & M19 & train & 16,384 & diagnostic & 201.87 & 246.34 & 22.03 & 236.56 & 17.19 & 0.10 \\
Mac Studio M4 Max & M20 & train & 0 & \textbf{primary} & 233.53 & 246.34 & 5.48 & 233.53 & 0.00 & 0.02 \\
Mac Studio M4 Max & M20 & train & 4,096 & diagnostic & 224.14 & 246.34 & 9.90 & 233.53 & 4.19 & 0.05 \\
Mac Studio M4 Max & M20 & train & 16,384 & diagnostic & 199.62 & 246.34 & 23.40 & 233.53 & 16.99 & 0.07 \\
Mac Studio M4 Max & M21 & train & 0 & \textbf{primary} & 246.34 & 246.34 & 0.00 & 246.34 & 0.00 & 0.25 \\
Mac Studio M4 Max & M21 & train & 4,096 & diagnostic & 236.32 & 246.34 & 4.24 & 246.34 & 4.24 & 0.09 \\
Mac Studio M4 Max & M21 & train & 16,384 & diagnostic & 209.32 & 246.34 & 17.68 & 246.34 & 17.68 & 0.04 \\
Mac Studio M4 Max & M22 & train & 0 & \textbf{primary} & 2142.88 & 2188.58 & 2.13 & 2019.85 & 5.74 & 0.26 \\
Mac Studio M4 Max & M22 & train & 4,096 & diagnostic & 1678.58 & 2188.58 & 30.38 & 2019.85 & 20.33 & 0.26 \\
Mac Studio M4 Max & M22 & train & 16,384 & diagnostic & 1019.72 & 2188.58 & 114.63 & 2019.85 & 98.08 & 0.10 \\
Mac Studio M4 Max & M23 & train & 0 & \textbf{primary} & 2237.19 & 2188.58 & 2.17 & 2237.19 & 0.00 & 0.31 \\
Mac Studio M4 Max & M23 & train & 4,096 & diagnostic & 1740.63 & 2188.58 & 25.73 & 2237.19 & 28.53 & 0.23 \\
Mac Studio M4 Max & M23 & train & 16,384 & diagnostic & 1040.47 & 2188.58 & 110.35 & 2237.19 & 115.02 & 0.13 \\
Mac Studio M4 Max & M03 & test & 0 & \textbf{primary} & 809.51 & 1178.24 & 45.55 & 1156.07 & 42.81 & 0.96 \\
Mac Studio M4 Max & M03 & test & 4,096 & diagnostic & 735.66 & 1178.24 & 60.16 & 1156.07 & 57.15 & 0.61 \\
Mac Studio M4 Max & M03 & test & 16,384 & diagnostic & 563.22 & 1178.24 & 109.20 & 1156.07 & 105.26 & 0.35 \\
Mac Studio M4 Max & M05 & test & 0 & \textbf{primary} & 4177.10 & 4696.60 & 12.44 & 4334.51 & 3.77 & 0.74 \\
Mac Studio M4 Max & M05 & test & 4,096 & diagnostic & 3275.07 & 4696.60 & 43.40 & 4334.51 & 32.35 & 0.84 \\
Mac Studio M4 Max & M05 & test & 16,384 & diagnostic & 1909.80 & 4696.60 & 145.92 & 4334.51 & 126.96 & 0.44 \\
Mac Studio M4 Max & M06 & test & 0 & \textbf{primary} & 253.94 & 241.61 & 4.85 & 232.03 & 8.63 & 0.10 \\
Mac Studio M4 Max & M06 & test & 4,096 & diagnostic & 241.32 & 241.61 & 0.12 & 232.03 & 3.85 & 0.10 \\
Mac Studio M4 Max & M06 & test & 16,384 & diagnostic & 227.12 & 241.61 & 6.38 & 232.03 & 2.16 & 0.08 \\
Mac Studio M4 Max & M07 & test & 0 & \textbf{primary} & 1277.14 & 1879.88 & 47.19 & 1734.94 & 35.85 & 0.19 \\
Mac Studio M4 Max & M07 & test & 4,096 & diagnostic & 1213.08 & 1879.88 & 54.97 & 1734.94 & 43.02 & 0.23 \\
Mac Studio M4 Max & M07 & test & 16,384 & diagnostic & 1058.33 & 1879.88 & 77.63 & 1734.94 & 63.93 & 0.18 \\
Mac Studio M4 Max & M08 & test & 0 & \textbf{primary} & 1301.39 & 1693.72 & 30.15 & 1563.14 & 20.11 & 0.34 \\
Mac Studio M4 Max & M08 & test & 4,096 & diagnostic & 1248.09 & 1693.72 & 35.71 & 1563.14 & 25.24 & 0.16 \\
Mac Studio M4 Max & M08 & test & 16,384 & diagnostic & 1099.00 & 1693.72 & 54.11 & 1563.14 & 42.23 & 0.09 \\
RTX 5080 & M01 & train & 0 & \textbf{primary} & 5624.55 & 4619.99 & 17.86 & 5624.55 & 0.00 & 5.17 \\
RTX 5080 & M01 & train & 4,096 & diagnostic & 5030.69 & 4619.99 & 8.16 & 5624.55 & 11.80 & 3.92 \\
RTX 5080 & M01 & train & 16,384 & diagnostic & 4336.80 & 4619.99 & 6.53 & 5624.55 & 29.69 & 2.60 \\
RTX 5080 & M04 & train & 0 & \textbf{primary} & 8633.84 & 13137.34 & 52.16 & 8633.84 & 0.00 & 3.75 \\
RTX 5080 & M04 & train & 4,096 & diagnostic & 8160.96 & 13137.34 & 60.98 & 8633.84 & 5.79 & 4.00 \\
RTX 5080 & M04 & train & 16,384 & diagnostic & 7086.04 & 13137.34 & 85.40 & 8633.84 & 21.84 & 2.75 \\
RTX 5080 & M09 & train & 0 & \textbf{primary} & 9911.27 & 13080.14 & 31.97 & 12128.70 & 22.37 & 5.67 \\
RTX 5080 & M09 & train & 4,096 & diagnostic & 9373.61 & 13080.14 & 39.54 & 12128.70 & 29.39 & 4.03 \\
RTX 5080 & M09 & train & 16,384 & diagnostic & 8143.16 & 13080.14 & 60.63 & 12128.70 & 48.94 & 3.62 \\
RTX 5080 & M10 & train & 0 & \textbf{primary} & 10379.35 & 13080.14 & 26.02 & 13011.54 & 25.36 & 5.85 \\
RTX 5080 & M10 & train & 4,096 & diagnostic & 9802.11 & 13080.14 & 33.44 & 13011.54 & 32.74 & 4.92 \\
RTX 5080 & M10 & train & 16,384 & diagnostic & 8448.91 & 13080.14 & 54.81 & 13011.54 & 54.00 & 3.69 \\
RTX 5080 & M11 & train & 0 & \textbf{primary} & 6452.64 & 6143.96 & 4.78 & 5697.05 & 11.71 & 3.37 \\
RTX 5080 & M11 & train & 4,096 & diagnostic & 6219.18 & 6143.96 & 1.21 & 5697.05 & 8.40 & 2.66 \\
RTX 5080 & M11 & train & 16,384 & diagnostic & 5738.49 & 6143.96 & 7.07 & 5697.05 & 0.72 & 2.17 \\
RTX 5080 & M12 & train & 0 & \textbf{primary} & 6861.31 & 6143.96 & 10.45 & 6111.74 & 10.92 & 4.42 \\
RTX 5080 & M12 & train & 4,096 & diagnostic & 6647.95 & 6143.96 & 7.58 & 6111.74 & 8.07 & 2.90 \\
RTX 5080 & M12 & train & 16,384 & diagnostic & 6079.76 & 6143.96 & 1.06 & 6111.74 & 0.53 & 2.86 \\
RTX 5080 & M15 & train & 0 & \textbf{primary} & 1725.38 & 2045.35 & 18.54 & 1725.38 & 0.00 & 3.23 \\
RTX 5080 & M15 & train & 4,096 & diagnostic & 1682.57 & 2045.35 & 21.56 & 1725.38 & 2.54 & 1.97 \\
RTX 5080 & M15 & train & 16,384 & diagnostic & 1538.25 & 2045.35 & 32.97 & 1725.38 & 12.17 & 2.17 \\
RTX 5080 & M16 & train & 0 & \textbf{primary} & 2024.12 & 2013.56 & 0.52 & 2024.12 & 0.00 & 3.50 \\
RTX 5080 & M16 & train & 4,096 & diagnostic & 1954.57 & 2013.56 & 3.02 & 2024.12 & 3.56 & 2.46 \\
RTX 5080 & M16 & train & 16,384 & diagnostic & 1771.53 & 2013.56 & 13.66 & 2024.12 & 14.26 & 2.22 \\
RTX 5080 & M17 & train & 0 & \textbf{primary} & 2043.07 & 2013.56 & 1.44 & 2043.07 & 0.00 & 3.11 \\
RTX 5080 & M17 & train & 4,096 & diagnostic & 1962.09 & 2013.56 & 2.62 & 2043.07 & 4.13 & 3.14 \\
RTX 5080 & M17 & train & 16,384 & diagnostic & 1783.83 & 2013.56 & 12.88 & 2043.07 & 14.53 & 2.23 \\
RTX 5080 & M18 & train & 0 & \textbf{primary} & 2042.02 & 2013.56 & 1.39 & 2042.02 & 0.00 & 17.04 \\
RTX 5080 & M18 & train & 4,096 & diagnostic & 2114.20 & 2013.56 & 4.76 & 2042.02 & 3.41 & 2.90 \\
RTX 5080 & M18 & train & 16,384 & diagnostic & 1907.04 & 2013.56 & 5.59 & 2042.02 & 7.08 & 2.39 \\
RTX 5080 & M22 & train & 0 & \textbf{primary} & 16588.19 & 17889.45 & 7.84 & 16588.19 & 0.00 & 7.30 \\
RTX 5080 & M22 & train & 4,096 & diagnostic & 14035.04 & 17889.45 & 27.46 & 16588.19 & 18.19 & 5.06 \\
RTX 5080 & M22 & train & 16,384 & diagnostic & 9948.68 & 17889.45 & 79.82 & 16588.19 & 66.74 & 3.46 \\
RTX 5080 & M23 & train & 0 & \textbf{primary} & 17795.63 & 17889.45 & 0.53 & 17795.63 & 0.00 & 8.37 \\
RTX 5080 & M23 & train & 4,096 & diagnostic & 14813.44 & 17889.45 & 20.76 & 17795.63 & 20.13 & 5.96 \\
RTX 5080 & M23 & train & 16,384 & diagnostic & 10330.65 & 17889.45 & 73.17 & 17795.63 & 72.26 & 3.83 \\
RTX 5080 & M05 & test & 0 & \textbf{primary} & 11706.91 & 38390.00 & 227.93 & 35597.53 & 204.07 & 39.19 \\
RTX 5080 & M05 & test & 4,096 & diagnostic & 13303.93 & 38390.00 & 188.56 & 35597.53 & 167.57 & 5.45 \\
RTX 5080 & M05 & test & 16,384 & diagnostic & 10873.99 & 38390.00 & 253.04 & 35597.53 & 227.36 & 4.90 \\
RTX 5080 & M08 & test & 0 & \textbf{primary} & 11433.58 & 13844.48 & 21.09 & 12837.44 & 12.28 & 4.86 \\
RTX 5080 & M08 & test & 4,096 & diagnostic & 10797.04 & 13844.48 & 28.22 & 12837.44 & 18.90 & 3.44 \\
RTX 5080 & M08 & test & 16,384 & diagnostic & 9215.27 & 13844.48 & 50.23 & 12837.44 & 39.31 & 3.36 \\
\end{longtable}
\endgroup
\end{landscape}
\section{Repeatability and calibration}
A separate repeatability series exists only for the MacBook Pro M4 Max; variability on the other hosts is represented by the within-invocation repetitions in the row tables, not by equivalent across-run series. The six Mac runs give mean decode throughput 166.457 tokens/s and sample between-run CV 1.613\%; prefill is 2133.091 tokens/s with CV 0.373\%. The max-minus-min spans are 4.34\% and 0.97\% of their respective means. Decode has a descriptive linear trend of 0.74\% per run. These are single-cell diagnostics, not general detection thresholds.
\begin{center}
\small
\begin{tabular}{rlrrrr}
\toprule
Run & Timestamp & Settle (s) & Prefill & Decode & Decode within-run SD\\
\midrule
1 & 2026-08-30T01:06:08.096930-07:00 & 45 & 2137.572 & 164.323 & 6.520 \\
2 & 2026-08-30T01:06:59.837356-07:00 & 45 & 2136.650 & 165.447 & 1.473 \\
3 & 2026-08-30T01:07:51.671530-07:00 & 45 & 2136.742 & 163.200 & 10.629 \\
4 & 2026-08-30T01:08:43.573411-07:00 & 45 & 2116.939 & 166.925 & 2.338 \\
5 & 2026-08-30T01:09:35.318847-07:00 & 45 & 2135.238 & 168.415 & 0.805 \\
6 & 2026-08-30T01:10:26.991582-07:00 & 45 & 2135.403 & 170.432 & 0.464 \\
\bottomrule
\end{tabular}
\end{center}
The append-only Mac measurement file also preserves a complete pre-protocol decode grid. It is excluded from every fit and score above. The comparison is MacBook-only and is not pooled with other hosts.
\begin{center}
\small
\begin{tabular}{lrrrrrrrr}
\toprule
Pass & All $n$ & Test $n$ & Test MAPE & Median & P90 & Max & All-row CV med. & All-row CV max\\
\midrule
Archived pre-protocol & 57 & 12 & 19.30 & 15.77 & 25.94 & 74.59 & 1.74 & 14.91 \\
Final Mac protocol-complete & 57 & 12 & 13.11 & 8.98 & 36.65 & 49.92 & 1.00 & 3.03 \\
\bottomrule
\end{tabular}
\end{center}
For comparison, the retained historical uncontrolled file contains two depth-0 observations of the same Qwen3.5-4B Q4\_K\_M cell:
\begin{center}
\small
\begin{tabular}{lrrr}
\toprule
Phase & Higher throughput & Lower throughput & Decrease (\%)\\
\midrule
prefill & 1467.424 & 1090.676 & 25.67 \\
decode & 103.616 & 87.594 & 15.46 \\
\bottomrule
\end{tabular}
\end{center}
\subsection{Host settings and calibration}
\begin{center}
\small
\begin{tabular}{lllrrrr}
\toprule
Host & Platform & Accelerator & Threads & Reps & Settle (s) & Requested ngl\\
\midrule
MacBook Pro M4 Max & Darwin-arm64 & arm (Metal) & 12 & 5 & 45 & 99 \\
Mac Studio M4 Max & Darwin-arm64 & arm (Metal) & 12 & 5 & 45 & 99 \\
RTX 5080 & Windows-AMD64 & NVIDIA GeForce RTX 5080 & 16 & 5 & 45 & 99 \\
\bottomrule
\end{tabular}
\end{center}
The requested \texttt{n\_gpu\_layers=99} setting asks the runtime to offload as many layers as it can; it is not direct telemetry proving that every tensor and KV allocation remained resident on the accelerator. No device-memory trace was recorded, so the appendix does not label these observations as confirmed fully resident.
\begin{center}
\small
\begin{tabular}{llrl}
\toprule
Host & Calibration & Value & Detail\\
\midrule
MacBook Pro M4 Max & CPU STREAM-style triad & 48.558 GB/s & 7 repetitions \\
MacBook Pro M4 Max & CPU FP32 matrix multiply & 2.458 TFLOP/s & 5 repetitions \\
MacBook Pro M4 Max & MPS device copy & 380.052 GB/s & arm (Metal) \\
MacBook Pro M4 Max & MPS FP16 matrix multiply & 13.950 TFLOP/s & arm (Metal) \\
MacBook Pro M4 Max & LLM reference ceiling & 377.244 GB/s & \nolinkurl{Qwen3.5-9B-Q8_0.gguf} \\
Mac Studio M4 Max & CPU STREAM-style triad & 77.670 GB/s & 7 repetitions \\
Mac Studio M4 Max & CPU FP32 matrix multiply & 3.017 TFLOP/s & 5 repetitions \\
Mac Studio M4 Max & MPS device copy & 396.727 GB/s & arm (Metal) \\
Mac Studio M4 Max & MPS FP16 matrix multiply & 15.001 TFLOP/s & arm (Metal) \\
Mac Studio M4 Max & LLM reference ceiling & 425.530 GB/s & \nolinkurl{Qwen3.8-27B-Q8_0.gguf} \\
RTX 5080 & CPU STREAM-style triad & 13.458 GB/s & 7 repetitions \\
RTX 5080 & CPU FP32 matrix multiply & 2.314 TFLOP/s & 5 repetitions \\
RTX 5080 & CUDA device copy & 801.061 GB/s & NVIDIA GeForce RTX 5080 \\
RTX 5080 & CUDA FP16 matrix multiply & 118.829 TFLOP/s & NVIDIA GeForce RTX 5080 \\
\bottomrule
\end{tabular}
\end{center}
For the fitted B1/B2 and P1/P2 models, the measured bandwidth or FLOP/s scale cancels algebraically after fitting on this same host. It sets the scale of B0 and supports roofline interpretation; it does not by itself produce the reported fitted-model accuracy. The exported row-level bandwidth field uses the device-copy result on all hosts. The LLM references are retained as diagnostics and are not the B0 bandwidth value used in the current exports.
\begin{center}
\small
\begin{tabular}{llrrrrr}
\toprule
Host & Probe & File GB & Decode & SD & CV (\%) & Effective GB/s\\
\midrule
MacBook Pro M4 Max & \nolinkurl{Qwen3.8-27B-Q8_0.gguf} & 29.05 & 7.100 & 4.199 & 59.14 & 196.653 \\
MacBook Pro M4 Max & \nolinkurl{Qwen3.5-9B-Q8_0.gguf} & 9.53 & 44.661 & 0.586 & 1.31 & 377.244 \\
MacBook Pro M4 Max & \nolinkurl{Qwen3.5-4B-Q8_0.gguf} & 4.48 & 70.817 & 5.524 & 7.80 & 317.432 \\
Mac Studio M4 Max & \nolinkurl{Qwen3.8-27B-Q8_0.gguf} & 29.05 & 15.364 & 0.080 & 0.52 & 425.530 \\
Mac Studio M4 Max & \nolinkurl{Qwen3.5-9B-Q8_0.gguf} & 9.53 & 48.773 & 0.159 & 0.33 & 411.974 \\
Mac Studio M4 Max & \nolinkurl{Qwen3.5-4B-Q8_0.gguf} & 4.48 & 79.528 & 0.146 & 0.18 & 356.475 \\
\bottomrule
\end{tabular}
\end{center}
\section{Output-projection sensitivity}
The additive output-projection term is reported as a sensitivity analysis. It is not selected: held-out improvement is not consistent across loss functions and hosts, and the term partly recharges weights already included in the streamed-byte term. Host rows and the unequally weighted all-host aggregate are shown separately. The one-term rows in this section are refit under the listed time-domain losses; they are not the median-ratio B2 headline and need not reproduce its MAPE.
\begin{center}
\small
\begin{tabular}{llllrrrrr}
\toprule
Host & Loss & Model & Split & $n$ & MAPE & Median & P90 & Max\\
\midrule
MacBook Pro M4 Max & absolute time & one term & train & 45 & 12.38 & 4.97 & 34.07 & 91.71 \\
MacBook Pro M4 Max & absolute time & one term & test & 12 & 19.26 & 14.43 & 46.77 & 61.03 \\
MacBook Pro M4 Max & absolute time & two terms & train & 45 & 10.78 & 4.47 & 32.04 & 77.90 \\
MacBook Pro M4 Max & absolute time & two terms & test & 12 & 38.56 & 35.19 & 71.12 & 82.26 \\
Mac Studio M4 Max & absolute time & one term & train & 45 & 15.32 & 8.29 & 38.34 & 96.54 \\
Mac Studio M4 Max & absolute time & one term & test & 15 & 20.84 & 15.60 & 47.57 & 68.22 \\
Mac Studio M4 Max & absolute time & two terms & train & 45 & 10.70 & 5.05 & 35.95 & 75.68 \\
Mac Studio M4 Max & absolute time & two terms & test & 15 & 38.17 & 32.69 & 72.58 & 81.63 \\
RTX 5080 & absolute time & one term & train & 36 & 14.18 & 11.77 & 28.14 & 45.36 \\
RTX 5080 & absolute time & one term & test & 6 & 44.01 & 31.40 & 97.73 & 105.71 \\
RTX 5080 & absolute time & two terms & train & 36 & 12.70 & 9.51 & 30.29 & 40.59 \\
RTX 5080 & absolute time & two terms & test & 6 & 26.08 & 24.58 & 51.38 & 51.91 \\
all hosts & absolute time & one term & train & 126 & 13.94 & 8.17 & 34.00 & 96.54 \\
all hosts & absolute time & one term & test & 33 & 24.48 & 15.60 & 60.69 & 105.71 \\
all hosts & absolute time & two terms & train & 126 & 11.30 & 5.53 & 33.61 & 77.90 \\
all hosts & absolute time & two terms & test & 33 & 36.11 & 32.69 & 71.30 & 82.26 \\
MacBook Pro M4 Max & relative time & one term & train & 45 & 11.40 & 4.35 & 28.13 & 83.22 \\
MacBook Pro M4 Max & relative time & one term & test & 12 & 15.17 & 11.87 & 40.27 & 53.89 \\
MacBook Pro M4 Max & relative time & two terms & train & 45 & 8.53 & 5.06 & 18.93 & 44.61 \\
MacBook Pro M4 Max & relative time & two terms & test & 12 & 22.20 & 20.37 & 40.79 & 48.48 \\
Mac Studio M4 Max & relative time & one term & train & 45 & 14.24 & 6.99 & 32.74 & 88.15 \\
Mac Studio M4 Max & relative time & one term & test & 15 & 17.47 & 14.72 & 41.28 & 61.04 \\
Mac Studio M4 Max & relative time & two terms & train & 45 & 9.62 & 4.87 & 20.97 & 49.20 \\
Mac Studio M4 Max & relative time & two terms & test & 15 & 24.98 & 24.90 & 42.73 & 48.71 \\
RTX 5080 & relative time & one term & train & 36 & 12.55 & 11.72 & 22.83 & 34.79 \\
RTX 5080 & relative time & one term & test & 6 & 40.81 & 29.12 & 83.35 & 90.75 \\
RTX 5080 & relative time & two terms & train & 36 & 11.74 & 10.36 & 21.21 & 28.73 \\
RTX 5080 & relative time & two terms & test & 6 & 27.64 & 23.25 & 52.96 & 55.20 \\
all hosts & relative time & one term & train & 126 & 12.75 & 8.34 & 28.06 & 88.15 \\
all hosts & relative time & one term & test & 33 & 20.88 & 12.31 & 53.36 & 90.75 \\
all hosts & relative time & two terms & train & 126 & 9.84 & 5.95 & 21.21 & 49.20 \\
all hosts & relative time & two terms & test & 33 & 24.45 & 23.26 & 47.47 & 55.20 \\
\bottomrule
\end{tabular}
\end{center}
\begin{landscape}
\section{Provenance and explicit limits}
The on-disk inputs used for this appendix have the following SHA-256 digests. The repository HEAD at generation was \texttt{28358bc8\allowbreak{}64720887\allowbreak{}cf9b6e14\allowbreak{}aebf8e1e\allowbreak{}3ff86f55}; the generator does not assert a clean working tree, so the file digests, rather than this commit alone, identify the inputs.
\begingroup
\small
\setlength{\tabcolsep}{3.5pt}
\renewcommand{\arraystretch}{1.08}
\begin{longtable}{@{}L{1.35in} L{2.55in} L{5.5in}@{}}
\caption{Exact inputs used to generate this appendix.}\label{tab:hashes}\\
\toprule
Role & Path & SHA-256 \\
\midrule
\endfirsthead
\caption[]{Exact inputs used to generate this appendix. (continued)}\\
\toprule
Role & Path & SHA-256 \\
\midrule
\endhead
\midrule
\multicolumn{3}{r}{Continued on next page}\\
\endfoot
\bottomrule
\endlastfoot
measurements (lun-mac) & \nolinkurl{results/measurements_lun-mac.csv} & \texttt{99dcbf9a\allowbreak{}bb945487\allowbreak{}77131989\allowbreak{}09703c1f\allowbreak{}77e00555\allowbreak{}f0258f7b\allowbreak{}f9d83e43\allowbreak{}95be99b4} \\
calibration (lun-mac) & \nolinkurl{results/calibration_lun-mac.json} & \texttt{7c374f21\allowbreak{}7b54540d\allowbreak{}eb3266af\allowbreak{}88ba4ba3\allowbreak{}93bfb85a\allowbreak{}c94e8714\allowbreak{}7a9c32ff\allowbreak{}de0696bc} \\
environment (lun-mac) & \nolinkurl{results/env_lun-mac.json} & \texttt{10ae6555\allowbreak{}45911c2b\allowbreak{}668dcd8b\allowbreak{}629effa2\allowbreak{}fd369170\allowbreak{}b14dbd38\allowbreak{}f1451d60\allowbreak{}94f6f4a6} \\
measurements (mac-studio-m4-max) & \nolinkurl{results/measurements_mac-studio-m4-max.csv} & \texttt{a9a77823\allowbreak{}7b2cc3cc\allowbreak{}297dd594\allowbreak{}0a930ece\allowbreak{}0cabd544\allowbreak{}5c1665b3\allowbreak{}de4c024a\allowbreak{}336bacf8} \\
calibration (mac-studio-m4-max) & \nolinkurl{results/calibration_mac-studio-m4-max.json} & \texttt{033df814\allowbreak{}3068779f\allowbreak{}7c6726e3\allowbreak{}bdff7b5d\allowbreak{}9f77a8fc\allowbreak{}ab3081b9\allowbreak{}200fd422\allowbreak{}ba3147fa} \\
environment (mac-studio-m4-max) & \nolinkurl{results/env_mac-studio-m4-max.json} & \texttt{ebeee6b0\allowbreak{}7a509c0e\allowbreak{}bba2dbd7\allowbreak{}938b9e58\allowbreak{}5496e65a\allowbreak{}35a76a50\allowbreak{}d4514be5\allowbreak{}c75ec9c8} \\
system profile (mac-studio-m4-max) & \nolinkurl{results/system_profile_mac-studio-m4-max.txt} & \texttt{e1ba22cb\allowbreak{}0ef3d3b7\allowbreak{}ce48fa13\allowbreak{}37e0c02e\allowbreak{}f9d7193a\allowbreak{}de113afa\allowbreak{}be997c79\allowbreak{}9d51ae88} \\
operating system (mac-studio-m4-max) & \nolinkurl{results/os_mac-studio-m4-max.txt} & \texttt{cd5d6399\allowbreak{}b3785b04\allowbreak{}ceaae750\allowbreak{}45a9ca9f\allowbreak{}80f0bbc3\allowbreak{}526eff9d\allowbreak{}0c36b170\allowbreak{}ed1ec348} \\
source commit (mac-studio-m4-max) & \nolinkurl{results/source_commit_mac-studio-m4-max.txt} & \texttt{f457f4d5\allowbreak{}0e3197d5\allowbreak{}11eefc85\allowbreak{}b20fcfe1\allowbreak{}1a8b8c25\allowbreak{}0924b4c4\allowbreak{}d89b56b4\allowbreak{}ba8bc6df} \\
measurement-source SHA-256 (mac-studio-m4-max) & \nolinkurl{results/measurement_source_tree_mac-studio-m4-max.sha256} & \texttt{7c15bd94\allowbreak{}f60ebdf3\allowbreak{}49cce6df\allowbreak{}465576fb\allowbreak{}4efb9dff\allowbreak{}098e8ffe\allowbreak{}4121ea72\allowbreak{}ed34dab0} \\
llama.cpp package (mac-studio-m4-max) & \nolinkurl{results/llama_cpp_package_mac-studio-m4-max.txt} & \texttt{61c81bc7\allowbreak{}6cf5830e\allowbreak{}c3477189\allowbreak{}ec72d3d2\allowbreak{}b48a76ea\allowbreak{}2f52d0fa\allowbreak{}2531ad3b\allowbreak{}2cd6b91e} \\
llama-bench SHA-256 (mac-studio-m4-max) & \nolinkurl{results/llama_bench_mac-studio-m4-max.sha256} & \texttt{6bf25f6a\allowbreak{}a7546aa4\allowbreak{}38b8d82d\allowbreak{}96f6dcf0\allowbreak{}2579ef22\allowbreak{}ceb14dac\allowbreak{}8e5a0d66\allowbreak{}6f04fe22} \\
model sources (mac-studio-m4-max) & \nolinkurl{results/model_sources_mac-studio-m4-max.json} & \texttt{aed45399\allowbreak{}09710d8d\allowbreak{}145fd7d3\allowbreak{}b42fcc37\allowbreak{}047413cc\allowbreak{}502abd77\allowbreak{}d78b8edb\allowbreak{}b10be342} \\
model integrity (mac-studio-m4-max) & \nolinkurl{results/model_integrity_mac-studio-m4-max.json} & \texttt{ab8a4993\allowbreak{}e5a3f341\allowbreak{}cec37124\allowbreak{}31dc9bd3\allowbreak{}51968c9d\allowbreak{}92aebf2b\allowbreak{}0fd95aef\allowbreak{}edaf7f4d} \\
Python packages (mac-studio-m4-max) & \nolinkurl{results/python_packages_mac-studio-m4-max.txt} & \texttt{480c7f98\allowbreak{}05b2d84e\allowbreak{}885c8b9b\allowbreak{}720cf538\allowbreak{}d0044990\allowbreak{}f04f76ab\allowbreak{}285b717c\allowbreak{}98f478f5} \\
Tectonic package (mac-studio-m4-max) & \nolinkurl{results/tectonic_package_mac-studio-m4-max.txt} & \texttt{e9945325\allowbreak{}0832d16d\allowbreak{}9a603ea0\allowbreak{}837f7fc3\allowbreak{}f7bcb540\allowbreak{}2927bbf7\allowbreak{}edb7048a\allowbreak{}15953e0b} \\
Tectonic SHA-256 (mac-studio-m4-max) & \nolinkurl{results/tectonic_mac-studio-m4-max.sha256} & \texttt{a6c2f0af\allowbreak{}071c38fc\allowbreak{}10f8b9ca\allowbreak{}56372535\allowbreak{}e4b63be5\allowbreak{}b36d76f5\allowbreak{}33bd6a9c\allowbreak{}ba8c9248} \\
measurements (rtx5080) & \nolinkurl{results/measurements_rtx5080.csv} & \texttt{d4373736\allowbreak{}f7a38bb9\allowbreak{}c80eeec7\allowbreak{}804681af\allowbreak{}073e01d7\allowbreak{}8ca9ff0b\allowbreak{}37e80b6f\allowbreak{}442deca4} \\
calibration (rtx5080) & \nolinkurl{results/calibration_rtx5080.json} & \texttt{c8dc32dc\allowbreak{}71f547c9\allowbreak{}a3f2afc7\allowbreak{}42f4342b\allowbreak{}9b229b6c\allowbreak{}b1466541\allowbreak{}9196a57d\allowbreak{}acb0032f} \\
environment (rtx5080) & \nolinkurl{results/env_rtx5080.json} & \texttt{bb4573ef\allowbreak{}46f13152\allowbreak{}005eab42\allowbreak{}22680682\allowbreak{}4d0873dc\allowbreak{}0439fdb4\allowbreak{}d283b381\allowbreak{}1074fbaa} \\
decode predictions & \nolinkurl{results/predictions.csv} & \texttt{6dd0b486\allowbreak{}fde80d56\allowbreak{}1731b179\allowbreak{}76864280\allowbreak{}bb24111a\allowbreak{}bea33f1b\allowbreak{}55ffedab\allowbreak{}6a863f79} \\
prefill predictions & \nolinkurl{results/predictions_prefill.csv} & \texttt{9f401bc7\allowbreak{}3378167c\allowbreak{}43a99cac\allowbreak{}a58e89b7\allowbreak{}a4b10da0\allowbreak{}e5adfbfa\allowbreak{}4e31ae33\allowbreak{}16aee310} \\
decode aggregate error table & \nolinkurl{results/error_table.csv} & \texttt{07b6e484\allowbreak{}7e4a3268\allowbreak{}ccbe6322\allowbreak{}65bd8a42\allowbreak{}70c0d459\allowbreak{}432226e6\allowbreak{}a38b3125\allowbreak{}72ef7612} \\
decode error table by host & \nolinkurl{results/error_table_by_host.csv} & \texttt{f5a23381\allowbreak{}35fb06bf\allowbreak{}339ee518\allowbreak{}0639a34c\allowbreak{}c34a0e86\allowbreak{}43ae3f3c\allowbreak{}35993c48\allowbreak{}27ffa5e2} \\
prefill aggregate error table & \nolinkurl{results/error_table_prefill.csv} & \texttt{460f1be7\allowbreak{}dfa5e642\allowbreak{}19f54405\allowbreak{}f8a40039\allowbreak{}dbad0238\allowbreak{}00ab003e\allowbreak{}cef67dd5\allowbreak{}3afb8786} \\
prefill error table by host & \nolinkurl{results/error_table_prefill_by_host.csv} & \texttt{b5b0e18f\allowbreak{}1896fd1f\allowbreak{}1b4c4266\allowbreak{}a29b9a19\allowbreak{}f271ad5c\allowbreak{}c46bb32e\allowbreak{}a18bd43a\allowbreak{}30308194} \\
manifest & \nolinkurl{results/model_manifest.json} & \texttt{2a672d62\allowbreak{}2a30fd21\allowbreak{}07a01735\allowbreak{}744de9ba\allowbreak{}24e97025\allowbreak{}c3b9b973\allowbreak{}b5df442b\allowbreak{}2b662ec6} \\
metadata & \nolinkurl{results/model_metadata.json} & \texttt{cb1afadb\allowbreak{}7fad3b6d\allowbreak{}9d2df88b\allowbreak{}7176d72f\allowbreak{}4c25068b\allowbreak{}bb865e8c\allowbreak{}9a33cc79\allowbreak{}a3e4a2e5} \\
repeatability (lun-mac) & \nolinkurl{results/repeatability_lun-mac.csv} & \texttt{f79f9565\allowbreak{}aebc662b\allowbreak{}de98d1fe\allowbreak{}fa44b29d\allowbreak{}73e83188\allowbreak{}c60b6bf6\allowbreak{}0a1061cc\allowbreak{}2de9f622} \\
archived uncontrolled comparison & \nolinkurl{results/contaminated/measurements_UNCONTROLLED_lun-mac.csv.bak} & \texttt{b344118d\allowbreak{}3902a9aa\allowbreak{}6530a611\allowbreak{}3a3c5d0c\allowbreak{}0310ae2a\allowbreak{}703ad498\allowbreak{}6c8146d7\allowbreak{}0821eb50} \\
analysis implementation & \nolinkurl{llmperf/analyze.py} & \texttt{7eefe014\allowbreak{}36ae78f1\allowbreak{}ebe32c39\allowbreak{}c236a6f0\allowbreak{}dc521e35\allowbreak{}b410145a\allowbreak{}e785f04f\allowbreak{}66e8e27c} \\
refinement implementation & \nolinkurl{llmperf/refine.py} & \texttt{d52acc54\allowbreak{}f91c253e\allowbreak{}1f05df31\allowbreak{}f5903e3c\allowbreak{}155584c0\allowbreak{}d1071643\allowbreak{}677ded87\allowbreak{}cf6b9e58} \\
metadata implementation & \nolinkurl{llmperf/common.py} & \texttt{21f8da71\allowbreak{}de805e7e\allowbreak{}ecbaa5dd\allowbreak{}9f5c8fc4\allowbreak{}5f5fb2e9\allowbreak{}0b67ce41\allowbreak{}bcfac8d3\allowbreak{}da18b973} \\
publication-figure wrapper & \nolinkurl{paper/icassp2027/generate_main_figures.py} & \texttt{4c0b7eca\allowbreak{}a86b523c\allowbreak{}70df76f3\allowbreak{}c79d943b\allowbreak{}b5dfb75a\allowbreak{}ba0117c9\allowbreak{}3d15433d\allowbreak{}7d390ee9} \\
appendix generator & \nolinkurl{paper/icassp2027/generate_supplement.py} & \texttt{f557811b\allowbreak{}c7d7084f\allowbreak{}d7acba1c\allowbreak{}3d61c236\allowbreak{}b7e40ec1\allowbreak{}dc4d1275\allowbreak{}7625c3da\allowbreak{}9817e934} \\
\end{longtable}
\endgroup
\end{landscape}
\begin{itemize}
\item MacBook Pro M4 Max, Mac Studio M4 Max, and RTX 5080 contributed unequal, overlapping cohorts. Headline coefficients are fit separately by measured host. Each leave-one-host-out audit fits on all remaining measured hosts and does not establish universal transfer to new runtime stacks or formats.
\item Every scored row records a requested \texttt{n\_gpu\_layers=99}, but no device-memory trace or runtime-reported resident-layer count was retained. This is a maximal-offload request, not proof of full accelerator residency; no partial-offload sweep or offload-cliff result exists.
\item The largest RTX modeled file-plus-KV footprint is 17.508 GB for \nolinkurl{Qwen3.8-27B-UD-Q3_K_XL.gguf} at depth 16,384, while calibration reports 17.094 GB of device memory. The differing accounting conventions further preclude a residency claim.
\item Runtime provenance is host-specific and is reproduced in the environment and auxiliary provenance files. Placeholder commits, project commits, and usage-text version output are not interpreted as immutable runtime-binary revisions.
\item The Studio acquisition records immutable repository revisions and per-file LFS SHA-256 identifiers for all 23 manifest entries. These identify the remote artifacts; they are not retrospective local-file hashes for the earlier MacBook and RTX copies.
\item Derived parameter counts and per-depth KV bytes are consumed from the frozen metadata snapshot during appendix generation. This reproduces the analysis but is not a fresh, independent metadata extraction.
\item Held-out coverage is small: 4 configurations on MacBook Pro M4 Max, 5 configurations on Mac Studio M4 Max, and 2 configurations on RTX 5080. Its observed quantization formats are MXFP4, Q4\_K, Q4\_K\_M. No learning curve establishes how many reference configurations suffice.
\item Decode MAPE treats three depths from each host--file configuration as observations; those rows are correlated. No narrow confidence claim is warranted.
\item Prefill's primary result is restricted to depth 0. Larger-depth predictions are scope diagnostics because the equation has no existing-prefix term. The retry rule is decode-only; the counts of prefill rows above each host's declared CV gate are 7 of 57 on MacBook Pro M4 Max, 0 of 60 on Mac Studio M4 Max, and 29 of 42 on RTX 5080.
\item Calibration-probe exclusions are host-specific (3 on MacBook Pro M4 Max, 3 on Mac Studio M4 Max, and 0 on RTX 5080). Failed configurations and probe identities are listed above; a recorded prompt-batch failure is not relabeled as an out-of-memory failure without supporting telemetry.
\item Qwen3.8-27B IQ2 has 64 layers and 26.90B parameters, whereas the other shown Qwen3.8-27B files have 65 layers and 27.32B parameters. The quantization plot is therefore a near-family comparison, not a controlled bit-format substitution for one identical network.
\end{itemize}
The metadata snapshot additionally records these parser/source Git object IDs:
\begin{center}
\small
\begin{tabular}{ll}
\toprule
Object & Identifier\\
\midrule
Code commit & \texttt{1bdce5e0\allowbreak{}d917dce1\allowbreak{}52cba480\allowbreak{}9d05fc19\allowbreak{}22a98797} \\
Measurement blob & \texttt{0ec72c61\allowbreak{}f2eaa371\allowbreak{}ba4f591a\allowbreak{}28964ec6\allowbreak{}02d0686b} \\
Predictions blob & \texttt{de86a428\allowbreak{}6026e3b8\allowbreak{}95b3c380\allowbreak{}9f9b4b0e\allowbreak{}b7dbda47} \\
Shape-log blob & \texttt{fa3d7b9c\allowbreak{}bf4a7312\allowbreak{}19c7d528\allowbreak{}29fde031\allowbreak{}c67e3a2d} \\
\bottomrule
\end{tabular}
\end{center}